\documentclass{article}
\usepackage[table]{xcolor}
\definecolor{RevisionBlue}{RGB}{0,0,0}
\newif\ifrevisionmarks
\revisionmarksfalse
\newcommand{\rev}[1]{\ifrevisionmarks{\color{RevisionBlue}#1}\else#1\fi}
\newenvironment{revision}{\begingroup\ifrevisionmarks\color{RevisionBlue}\fi}{\endgroup}
\usepackage{iclr2027_conference,times}

\usepackage{amsmath,amsfonts,bm}

\def\eqref#1{equation~\ref{#1}}

\def\1{\bm{1}}

\DeclareMathAlphabet{\mathsfit}{\encodingdefault}{\sfdefault}{m}{sl}
\SetMathAlphabet{\mathsfit}{bold}{\encodingdefault}{\sfdefault}{bx}{n}

\newcommand{\E}{\mathbb{E}}

\newcommand{\R}{\mathbb{R}}

\DeclareMathOperator*{\argmin}{arg\,min}

\usepackage{hyperref}
\hypersetup{hidelinks,hypertexnames=false}
\usepackage{url}
\usepackage{booktabs}
\usepackage{colortbl,array}
\usepackage{float}
\usepackage{placeins}
\usepackage{amsmath,amssymb,amsthm}
\usepackage{mathtools}
\usepackage{graphicx}
\usepackage{algorithm}
\usepackage{algpseudocode}

\newtheorem{proposition}{Proposition}
\newtheorem{theorem}{Theorem}
\newtheorem{lemma}{Lemma}
\newtheorem{corollary}{Corollary}
\newtheorem{assumption}{Assumption}

\definecolor{PAOSAccentBlue}{RGB}{224,241,247}
\newcommand{\simplex}{\Delta_S}

\providecommand{\cX}{\mathcal{X}}
\providecommand{\cY}{\mathcal{Y}}

\providecommand{\cN}{\mathcal{N}}
\providecommand{\bF}{\mathbf{F}}
\providecommand{\bH}{\mathbf{H}}
\providecommand{\bPhi}{\boldsymbol{\Phi}}
\providecommand{\balpha}{\boldsymbol{\alpha}}
\providecommand{\blambda}{\boldsymbol{\lambda}}
\providecommand{\br}{\mathbf{r}}
\providecommand{\bq}{\mathbf{q}}
\providecommand{\bx}{\mathbf{x}}
\providecommand{\by}{\mathbf{y}}
\providecommand{\bv}{\mathbf{v}}
\providecommand{\bd}{\mathbf{d}}
\providecommand{\ba}{\mathbf{a}}
\providecommand{\be}{\mathbf{e}}
\providecommand{\bh}{\mathbf{h}}
\providecommand{\bw}{\mathbf{w}}
\providecommand{\bz}{\mathbf{z}}
\providecommand{\bxi}{\boldsymbol{\xi}}
\providecommand{\one}{\mathbf{1}}

\renewcommand{\eqref}[1]{(\ref{#1})}

\title{Closing the Pareto-Response Loop:\\
Preference-Aware Optimistic Selection for\\
Multi-Objective Bilevel Learning}
\author{Jiaxin Gao\\
The Hong Kong Polytechnic University\\
\texttt{jiaxinn.gao@outlook.com}
\And
Zhenyu Liang\\
The Hong Kong Polytechnic University\\
\texttt{zhenyu.liang@connect.polyu.hk}
\AND
Yaohua Liu\\
The University of Hong Kong\\
\texttt{liuyaohua.918@gmail.com}
\And
Ran Cheng\\
The Hong Kong Polytechnic University\\
\texttt{ranchengcn@gmail.com}}
\iclrfinalcopy

\begin{document}
\maketitle

\begin{abstract}
\rev{Advanced learning tasks increasingly couple lower-level model training with
hyper-training over multiple conflicting upper-level objectives.}
Multi-Objective Bilevel Learning (MOBL) models this hierarchical coupling,
with the learned lower-level response shaping the Pareto geometry available
to upper-level decisions.
When lower-level optimization is non-convex or finite-step, response
formation can be \emph{highly path-dependent}.
Existing response-first pipelines leave response formation open-loop because
Pareto feedback does not directly guide lower-level training.
We introduce \textbf{Preference-Aware Optimistic Selection (PAOS)}, which
closes this loop by conditioning response selection on the Pareto geometry
induced by the current response.
A stateful selector accumulates preference-weighted signals from successive
Pareto subproblems and uses them to construct the lower objective for the
next training response.
On an analytically tractable double-well problem, PAOS systematically selects
the preference-compatible branch from near-boundary starts.
Across data hyper-cleaning, few-shot bilevel learning, physics-informed
neural networks, and medical image segmentation, PAOS achieves the strongest
performance among the evaluated MOBL methods.
These results demonstrate the value of closed-loop response formation in MOBL.
\end{abstract}

\section{Introduction}
\label{sec:introduction}

\rev{In recent years, a growing number of advanced learning tasks have involved
coupling lower-level training with hyper-training over multiple conflicting
upper-level objectives.
This coupling is naturally formulated as Multi-Objective Bilevel Learning
(MOBL)}, in which the upper-level criteria are evaluated through a response
produced by lower-level optimization
\citep{zhang2026mobl,chauhan2026review,gu2023morbit,wang2024conditional,yang2026bilevel}.
The learned response therefore determines not only task performance but also
the trade-offs available to upper-level decision-making.
Such couplings arise in data reweighting, meta-learning, and scientific
machine learning
\citep{franceschi2018bilevel,dagreou2022framework}.
\rev{In image-classification data hyper-cleaning, shared sample weights shape
the training of classifiers under different label-corruption regimes; the
resulting clean-validation losses define the competing upper-level objectives.}
Coordinating these objectives therefore depends on how the underlying
responses are formed, making response formation central to preference-aware
MOBL.

Existing approaches provide increasingly effective tools for coordinating
conflicting objectives.
Preference-guided multi-objective learning targets user-specified Pareto
regions
\citep{mahapatra2020multi,navon2021learning,chen2024ferero,chen2025pareto}.
MOBL methods further develop multi-gradient aggregation, robust formulations,
first-order reformulations, Pareto-stationary algorithms, and
preference-aware subproblems
\citep{ye2021moml,gu2023morbit,ye2024forum,yang2024gmoba,zhang2026mobl}.
Response-first multi-hypergradient methods coordinate objectives
after the lower solver has formed a response.
Their Pareto decisions influence subsequent responses through upper-variable
updates, but the resulting mixture does not directly enter the lower
training objective.
We use \emph{open-loop response formation} to denote this missing direct
channel between Pareto coordination and lower-level training.

This distinction matters when lower-level optimization is non-convex or
performed for a finite number of steps.
The reached response can be \emph{highly path-dependent}, reflecting
initialization, optimization trajectories, and transient training dynamics
\citep{liu2020bda,vicol2022implicit,arbel2022selection,liu2023bamm}.
Different responses induce different upper-level objectives and
hypergradients, changing the Pareto geometry on which coordination operates.
The response is therefore not merely an outcome to evaluate:
its formation offers a means of shaping subsequent trade-offs.
This observation raises the central question:
\textbf{Can the upper-level preference actively select the lower-level
response from which the next Pareto geometry emerges?}
We address this question with
\textbf{Preference-Aware Optimistic Selection (PAOS)}.

\begin{figure}[t]
	\centering
	\includegraphics[width=\linewidth]{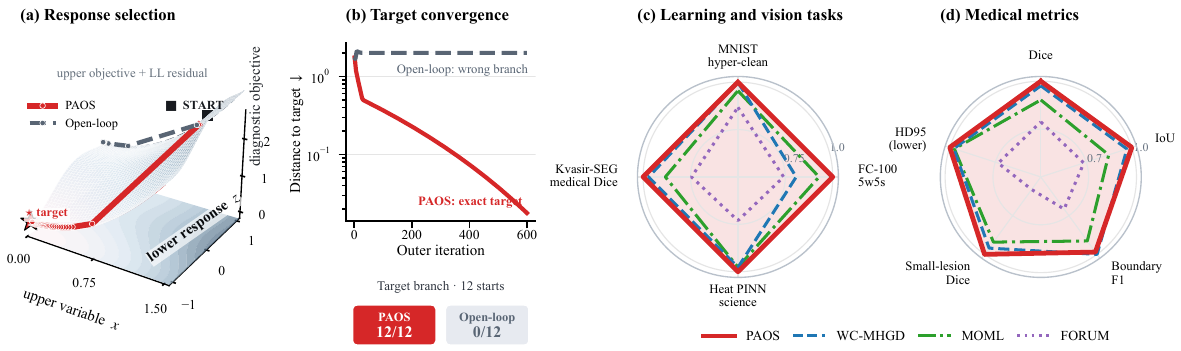}
	\vspace{-0.7cm}
	\caption{
		\textbf{Response formation matters for MOBL.}
		\textbf{(a--b)} Pareto feedback steers PAOS toward the preferred branch
		in all 12 near-boundary double-well runs, while open-loop optimization
		follows the incompatible branch (Appendix~\ref{app:dw_aligned}).
		\textbf{(c--d)} The remaining panels preview comparative results across
		learning benchmarks and medical segmentation metrics.
	}
	\label{fig:motivation}
\end{figure}

PAOS makes the response-selection criterion depend on the Pareto
geometry induced by the current response.
It combines the user preference with the current Pareto mixture and
integrates this signal into a stateful selector.
The selector controls a selected lower-level objective that guides
subsequent response formation.
The next response, in turn, induces the geometry used for the next Pareto
decision, closing the Pareto-response loop.
This mechanism realizes preference-aware optimistic branch selection when
the lower solution set is non-singleton, and preference-guided finite-budget
response shaping when the lower response is unique.

Figure~\ref{fig:motivation} previews its empirical benefits across data
hyper-cleaning \citep{ren2018reweight,franceschi2018bilevel},
few-shot bilevel learning
\citep{oreshkin2018tadam,rajeswaran2019implicit},
physics-informed neural networks \citep{raissi2019physics},
and medical image segmentation \citep{jha2020kvasir}.
Analytical experiments demonstrate preference-consistent branch selection,
while controlled ablations establish cumulative gains from response
selection, preference conditioning, and dynamic Pareto feedback.
Together, these results position response formation as an actionable
degree of freedom in MOBL.
Our contributions are threefold:
\begin{itemize}
    \item \textbf{A closed-loop perspective on MOBL.}
    We identify a direct feedback channel through which Pareto coordination
    can shape lower-level response formation and, consequently, the
    geometry of subsequent upper-level trade-offs.

    \item \textbf{Preference-Aware Optimistic Selection.}
    We develop a stateful selector that converts preference-weighted Pareto
    mixtures into a controlled lower-level objective, enabling optimistic
    branch selection and finite-budget response shaping.

    \item \textbf{Mechanistic and cross-domain validation.}
    We demonstrate preference-consistent branch selection on an analytic
    problem and leading performance across four domains, with controlled
    ablations isolating dynamic-feedback gains beyond static preference
    selection.
\end{itemize}
\section{Related Work}
\label{sec:related_work}

\paragraph{Bilevel Optimization and Learning.}
Bilevel optimization provides a common formulation for learning through
nested optimization, including hyperparameter optimization and meta-learning
\citep{liu2022bilevelsurvey}.
Hierarchical formulations also appear in image enhancement
\citep{gao2023hyperguidance,gao2024collaba,xie2024water,gao2024trilevel},
adversarial transfer \citep{liu2026bmat,liu2026mabt}, constraint learning
\citep{liu2024constraint}, and automated optimizer design
\citep{yu2026autopso}.
Recent advances improve computational efficiency through variance-reduced
hypergradient estimation \citep{dagreou2022framework,dagreou2024lower}
and fully first-order algorithms \citep{chen2025nearoptimal}.
Beyond computational efficiency, response-selection studies examine how
initialization and optimization dynamics determine the attained lower-level
solution \citep{vicol2022implicit,arbel2022selection,masiha2026select}.
\rev{Bilevel Descent Aggregation (BDA) incorporates upper-level information into
lower-level optimization to realize optimistic selection
\citep{liu2020bda,liu2023descent}.}
\rev{Whereas BDA guides selection through a prescribed scalar upper
criterion, including a fixed scalarization, PAOS derives the selection
signal from the current multi-objective Pareto subproblem. Selection thus
adapts to the evolving response-conditioned objective and hypergradient
geometry.}

\paragraph{Multi-Objective and Preference-Aware Learning.}
Multi-objective learning coordinates conflicting criteria through gradient
aggregation and preference-guided optimization. Recent methods address
stochastic gradient bias \citep{fernando2023mitigating} and construct
conflict-free updates for competing physics-informed losses
\citep{liu2025config}.
Preference-aware approaches make the desired trade-off explicit:
FERERO accommodates relative and absolute preferences through constrained
vector optimization \citep{chen2024ferero}, while optimization over the
Pareto set casts preference guidance as a semivectorial bilevel problem
\citep{chen2025pareto}.
These developments provide complementary mechanisms for coordinating
objectives and identifying preferred trade-offs.
\rev{PAOS uses the resulting preference-conditioned Pareto mixture to
control lower-level response formation, connecting objective-space
coordination to the training process that generates the next geometry.}

\paragraph{Multi-Objective Bilevel Optimization and Learning.}
Multi-objective bilevel optimization encompasses evolutionary search and
gradient-based learning. Evolutionary approaches model lower-level solution
mappings with conditional generative networks
\citep{wang2024conditional} or adapt lower-level search through multiple
search modes \citep{yang2026bilevel}.
Among gradient-based methods, MOML~\citep{ye2021moml} alternates lower-level
gradient optimization with upper-level multi-objective updates.
gMOBA~\citep{yang2024gmoba} iteratively approximates lower-level solutions
and implicit hypergradients, coupled with a proximal multi-objective
upper update.
FORUM~\citep{ye2024forum} reformulates lower-level optimality as a
value-function constraint and solves the resulting constrained
multi-objective problem through first-order multi-gradient aggregation.
Hypergradient-based coordination includes MGDA-HG, which applies
minimum-norm gradient aggregation to upper-level hypergradients
\citep{sener2018multi,zhang2026mobl}, and
WC-MHGD~\citep{zhang2026mobl}, which combines multi-hypergradient descent
with weighted-Chebyshev scalarization for preference-guided Pareto
front exploration.
Related developments include MORBiT's min-max formulation for robust
bilevel learning \citep{gu2023morbit}, non-convex lower-level optimization,
and equality-constrained reformulations
\citep{jiang2026efficient,zhang2026tale}.
\rev{PAOS connects Pareto coordination to response selection: the current
preference-conditioned Pareto mixture updates a stateful selector that
shapes the next lower-level solve, whose response in turn determines the
subsequent Pareto geometry.}

\section{Method}
\label{sec:method}

\begin{figure*}[t]
    \centering
    \includegraphics[width=\textwidth]{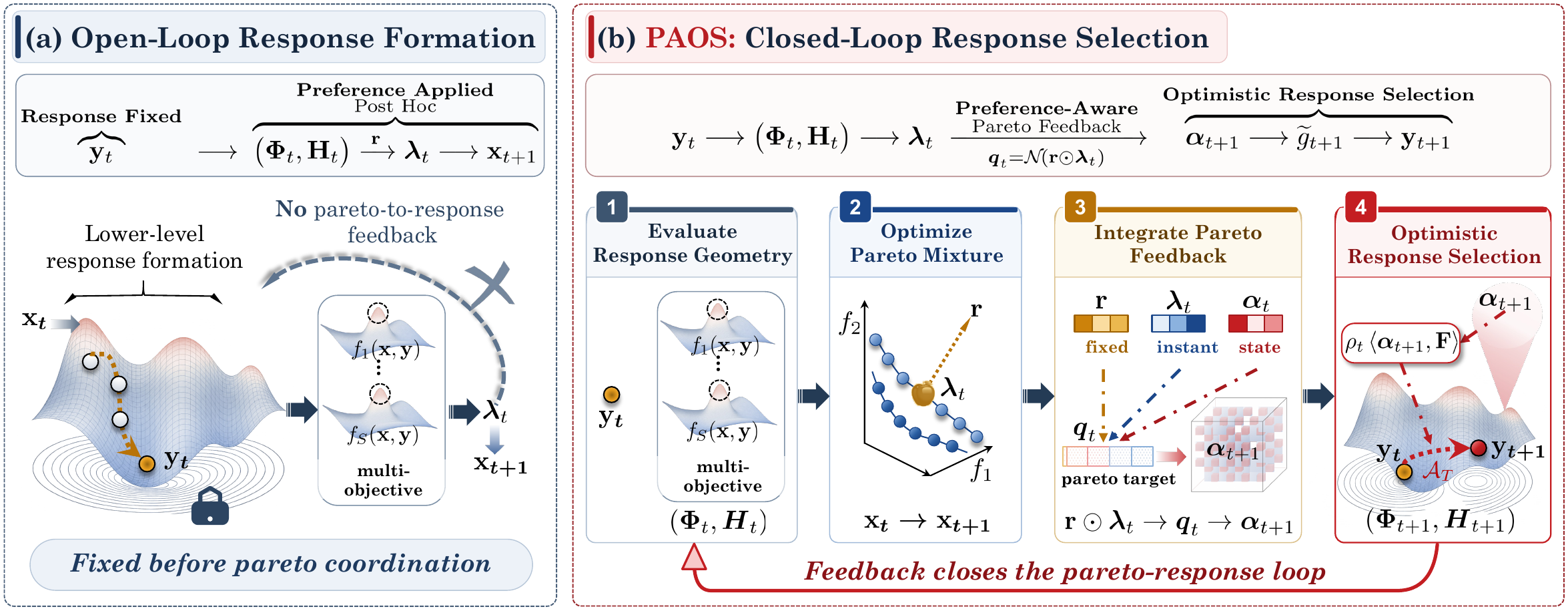}
    \vspace{-0.7cm}
    \caption{
\textbf{Closing the Pareto-response loop.}
(a) In the reference open-loop pipeline, lower-level training forms a response
before Pareto coordination.
(b) In PAOS, the current response informs a preference-conditioned Pareto
decision, which in turn guides the next response. This recurring dependence
closes the loop between training and Pareto coordination.
    }
    \label{fig:paos_overview}
\end{figure*}

\rev{PAOS couples the selected response map to its induced Pareto geometry:
the selector shapes the response, whose Pareto mixture updates the next selector.}
Figure~\ref{fig:paos_overview} summarizes this interaction.
Sections~\ref{sec:preliminaries}--\ref{sec:theory_main} develop the
formulation, selection, feedback, and theory;
Algorithm~\ref{alg:paos} gives the complete procedure.

\subsection{Preliminaries}
\label{sec:preliminaries}
\label{sec:open_loop}

\paragraph{Multi-objective bilevel optimization.}
We study multi-objective bilevel optimization for machine learning
training with a scalar lower-level objective.
Let $\bx\in\cX\subseteq\R^p$ and $\by\in\cY\subseteq\R^q$ denote
\rev{upper-level learning controls (i.e., hyper-training variables) and
lower-level model parameters (i.e., training variables), respectively.}
The $S$ upper-level objectives are collected in
$\bF(\bx,\by):=(f_1(\bx,\by),\ldots,f_S(\bx,\by))^\top$
and coupled through the lower-level training objective $g$.
For a selected response branch, MOBL seeks a Pareto solution of
\begin{equation}
\label{eq:mobl}
\min_{\bx\in\cX}\quad
\bPhi(\bx):=\bF\bigl(\bx,\by^\star(\bx)\bigr), \quad
\by^\star(\bx)\in\cY^\star(\bx)
:=\argmin_{\by\in\cY}g(\bx,\by).
\end{equation}
Here $\bPhi:\cX\to\R^S$ is the upper-level objective map induced by
the selected branch.
\rev{In the learning tasks below, the upper (hyper-training) level chooses
data weights, an encoder, or loss weights through $\bx$. Lower-level training
produces model parameters $\by$, idealized by $\by^\star(\bx)$ in
Eq.~\eqref{eq:mobl}, whose performance determines the upper-level trade-off.}
A preference $\br\in\operatorname{int}(\simplex)$ specifies the desired
trade-off, where
$\simplex:=\{\bw\in\R_+^S:\one^\top\bw=1\}$.

\paragraph{Open-loop response formation.}
Finite-step lower solvers produce responses that depend on
initialization, local geometry, and the optimization path.
In our reference pipeline, the lower solver operates on $g$, and the
resulting response is used for preference-conditioned Pareto coordination
(Figure~\ref{fig:paos_overview}(a)).
The Pareto decision updates the upper variable, which shapes subsequent
lower-level responses.
We call this organization \emph{open-loop response formation}.

\subsection{Preference-Aware Optimistic Selection}
\label{sec:selected_response}

\paragraph{Optimistic response selection.}
When the lower-level solution set contains multiple responses, selecting
among them can change the induced upper-level trade-off.
For a mixture $\balpha\in\simplex$, let
$\psi_{\balpha}(\bx,\by):=\langle\balpha,\bF(\bx,\by)\rangle$
denote the scalarized upper-level criterion.
We define the ideal optimistic response set as
\begin{equation} 
\label{eq:ideal_optimistic}
    \cY^{\mathrm{opt}}_{\balpha}(\bx)
    :=\argmin_{\by\in\cY^\star(\bx)}\psi_{\balpha}(\bx,\by).
\end{equation}
The lower-level objective $g$ determines the admissible response set
$\cY^\star(\bx)$, and $\psi_{\balpha}$ selects its minimizers within
that set.
This extends optimistic bilevel selection \citep{liu2020bda} to a
mixture of upper-level objectives.

\paragraph{Stateful selection surrogate.}
To incorporate the upper-level trade-off into response formation,
PAOS augments the lower objective with an upper-objective mixture
controlled by a stateful selector.
At iteration $t$, it forms the selected objective and warm-starts a
$T$-step lower solve (Step~4 of Figure~\ref{fig:paos_overview}(b)):
{\small
\begin{equation}
\label{eq:selected_lower}
\widetilde g_t(\bx,\by)
:=g(\bx,\by)+\rho_t\langle\balpha_t,\bF(\bx,\by)\rangle =g(\bx,\by)+\rho_t\sum_{s=1}^{S}\alpha_{t,s}f_s(\bx,\by), \quad \by_t:=\mathcal A_T(\by_{t-1};\bx_t,\widetilde g_t).
\end{equation}
}
Here $\rho_t\geq0$ controls the selection strength,
$\balpha_t\in\simplex$ is the response-selector state, and
$\mathcal A_T$ denotes $T$ lower-level optimization steps initialized
at $\by_{t-1}$.
The upper state $\bx_t$ and selector $\balpha_t$ remain fixed
throughout the lower solve.
We initialize $\balpha_0=\br$.
The user preference specifies the target trade-off, while subsequent
Pareto mixtures update $\balpha_t$ as the response-conditioned
geometry evolves.
The selected response induces the geometry used for Pareto coordination,
whose mixture then informs the next selector update.

\rev{For fixed $\bx$ and $\balpha$, under
Assumption~\ref{assump:existence},
Proposition~\ref{prop:optimistic_consistency} establishes that every
accumulation point of exact surrogate minimizers as $\rho\downarrow0$
belongs to the optimistic set in Eq.~\eqref{eq:ideal_optimistic}.}
With a unique lower-level response, the selector guides the
finite-step optimization trajectory through the perturbed objective.

\emph{Appendix~\ref{app:optimistic_proof} proves the small-$\rho$ selection
limit, while Appendix~\ref{app:local_perturbation} quantifies the induced
response and Pareto-geometry shifts.}

\subsection{Closed-Loop Pareto-Response Optimization}
\label{sec:closed_loop}

\paragraph{Response-conditioned Pareto geometry.}
We compute hypergradients through the selected response map and use
weighted-Chebyshev coordination to combine them.
At the selected response $(\bx_t,\by_t)$, PAOS evaluates
$\bPhi_t:=\bF(\bx_t,\by_t)$ and differentiates each objective through the
selected lower problem (Step~1 of Figure~\ref{fig:paos_overview}(b)). All
derivatives below are evaluated at $(\bx_t,\by_t)$ while
$\balpha_t$ and $\rho_t$ are held fixed. Define the damped selected-response
Hessian $A_t:=\nabla_{\by\by}^2\widetilde g_t+\delta I\in\R^{q\times q}$
and mixed derivative matrix
$B_t:=\nabla_{\bx\by}^2\widetilde g_t\in\R^{p\times q}$, where $\delta>0$.
For each objective
$s\in[S]$, warm-started conjugate gradients produce $\bv_{t,s}$ satisfying
$A_t\bv_{t,s}\approx\nabla_{\by}f_s$. The resulting implicit
hypergradient and hypergradient matrix are
\begin{equation}
\label{eq:hypergradient}
    \widehat{\bh}_{t,s}:=\nabla_{\bx}f_s-B_t\bv_{t,s},
    \qquad
    \bH_t:=[\widehat{\bh}_{t,1},\ldots,\widehat{\bh}_{t,S}]
    \in\R^{p\times S}.
\end{equation}
\emph{Finite-response, CG, and damping errors are analyzed in
Appendix~\ref{app:hg_error}.}

\paragraph{Preference-aware Pareto feedback.}
Stages~2-3 of Figure~\ref{fig:paos_overview}(b) convert the
response-conditioned geometry into the control signal that forms the next
lower-level response. Given $(\bPhi_t,\bH_t)$, PAOS computes an
instantaneous Pareto mixture through the weighted-Chebyshev
multi-hypergradient subproblem:
\begin{equation}
\label{eq:wc_qp}
    \blambda_t\in\argmin_{\blambda\in\simplex}
    \|\bH_t(\br\odot\blambda)\|_2^2
    -u\,\blambda^\top(\br\odot\bPhi_t).
\end{equation}
Here $\blambda_t\in\simplex$ reconciles the current objective conflicts,
$\br\in\operatorname{int}(\simplex)$ is the fixed user preference,
$\odot$ denotes element-wise multiplication, and $u\geq0$ weights the
preference-aware objective-space term. The effective mixture
$\br\odot\blambda_t$ determines the upper-level direction and update:
\begin{equation}
\label{eq:upper_update}
    \bd_t:=\bH_t(\br\odot\blambda_t),
    \qquad
    \bx_{t+1}:=\Pi_{\cX}(\bx_t-\eta_x\bd_t).
\end{equation}
Here $\eta_x>0$ is the upper-level step size, $\Pi_{\cX}$ is the
projection onto $\cX$ (the identity when $\cX=\R^p$).

The defining PAOS operation is to reuse the same Pareto decision as feedback
for subsequent response formation. Let
$\cN(\ba):=\ba/(\one^\top\ba)$ denote simplex normalization. PAOS forms the
instantaneous Pareto target and updates its response-selector state as
\begin{equation}
\label{eq:alpha_update}
    \bq_t:=\cN(\br\odot\blambda_t),
    \qquad
    \balpha_{t+1}:=(1-\beta)\balpha_t+\beta\bq_t,
    \quad \beta\in[0,1].
\end{equation}
The four quantities have distinct roles: $\br$ specifies the global intent,
$\blambda_t$ resolves the current response-conditioned geometry, $\bq_t$ is
the resulting instantaneous target, and $\balpha_{t+1}$ carries that target
into the next selected lower objective. The feedback rate $\beta$ controls
how rapidly current Pareto decisions enter response selection: $\beta=0$
recovers static preference selection when $\balpha_0=\br$, whereas $\beta>0$
closes the feedback loop. With $\br\in\operatorname{int}(\simplex)$,
the normalization is well-defined, and the convex-combination update
preserves $\balpha_{t+1}\in\simplex$.

\paragraph{Open- versus closed-loop recursion.}
Figure~\ref{fig:paos_overview} contrasts the reference response-first path
with PAOS. Both update $\bx$ using the Pareto mixture;
PAOS additionally routes that mixture through $\bq_t$ and $\balpha_{t+1}$,
directly shaping the next lower solve:
{\small
\begin{equation}
	\label{eq:closed_loop}
	\resizebox{\linewidth}{!}{$\displaystyle
		{\color{black}
			\text{\bfseries Open loop:}}\quad
			\overbrace{\by_t}^{
				\substack{
					\text{\scriptsize Response}\\[2pt]
					\text{\scriptsize\bfseries Fixed}
			}}
			\rightarrow
			\overbrace{
				(\bPhi_t,\bH_t)
				\xrightarrow{\,\br\,}
				\blambda_t
				\rightarrow
				\bx_{t+1}
			}^{
				{\color{black}\text{\scriptsize\bfseries Preference Applied Post Hoc}}
		}
		\quad\Big|\quad
		{\color{black}
			\text{\bfseries PAOS:}}\quad
			\by_t
			\rightarrow
			(\bPhi_t,\bH_t)
			\rightarrow
			\blambda_t
			\xrightarrow[
			\bq_t=\cN(\br\odot\blambda_t)
			]{
				\substack{
					\text{\scriptsize\bfseries Preference-Aware}\\[2pt]
					\text{\scriptsize Pareto Feedback}
			}}
			\overbrace{
				\balpha_{t+1}
				\rightarrow
				\widetilde g_{t+1}
				\rightarrow
				\by_{t+1}
			}^{
				{\color{black}\text{\scriptsize\bfseries Optimistic Response Selection}}
		}
		$}
\end{equation}
}
Thus, PAOS turns the current response into a source of Pareto feedback that
actively shapes the response formed at the next iteration.

\begin{algorithm}[t]
\caption{\textbf{Preference-Aware Optimistic Selection (PAOS)}}
\label{alg:paos}
\begin{algorithmic}[1]
\Require Initial states $\bx_0,\by_{-1}$; preference $\br$; outer, lower,
and CG steps $N,T,K$
\Require Learning rates $\eta_x,\eta_y$; feedback rate $\beta$; damping
$\delta$; selection schedule $\{\rho_t\}_{t=0}^{N-1}$
\State $\balpha_0\gets\br$; initialize CG states
$\{\bv_{-1,s}\}_{s=1}^{S}$
\For{$t=0,\ldots,N-1$}
    \Statex \textcolor{black}{\(\triangleright\)
    \textbf{Selected-response formation}}
    \State {\color{black}
    $\widetilde g_t\gets g+\rho_t\langle\balpha_t,\bF\rangle$;
    $\by_t^{(0)}\gets\by_{t-1}$}
    \For{$j=0,\ldots,T-1$}
        \State $\by_t^{(j+1)}\gets\by_t^{(j)}
        -\eta_y\nabla_{\by}\widetilde g_t(\bx_t,\by_t^{(j)})$
    \EndFor
    \State $\by_t\gets\by_t^{(T)}$; evaluate $\bPhi_t$
    \Statex \textcolor{black}{\(\triangleright\)
    \textbf{Response-conditioned geometry}}
    \For{$s=1,\ldots,S$}
        \State Approximately solve
        $(\nabla_{\by\by}^2\widetilde g_t+\delta I)\bv_{t,s}
        =\nabla_{\by}f_s$ using $K$ warm-started CG steps
        \State $\widehat{\bh}_{t,s}\gets
        \nabla_{\bx}f_s-\nabla_{\bx\by}^2\widetilde g_t\bv_{t,s}$
    \EndFor
    \State $\bH_t\gets
    [\widehat{\bh}_{t,1},\ldots,\widehat{\bh}_{t,S}]$;
    solve Eq.~\eqref{eq:wc_qp} for $\blambda_t$
    \State $\bd_t\gets\bH_t(\br\odot\blambda_t)$;
    $\bx_{t+1}\gets\Pi_{\cX}(\bx_t-\eta_x\bd_t)$
    \Statex \textcolor{black}{\(\triangleright\)
    \textbf{Preference-aware Pareto feedback}}
    \State {\color{black}
    $\bq_t\gets\cN(\br\odot\blambda_t)$}, {\color{black}
    $\balpha_{t+1}\gets(1-\beta)\balpha_t+\beta\bq_t$}
\EndFor
\State \Return the evaluated trajectory and selector state
$\{(\bx_t,\by_t)\}_{t=0}^{N-1},\balpha_N$
\end{algorithmic}
\end{algorithm}

\emph{Appendix~\ref{app:selector_schedule} specifies the selection schedule;
Appendices~\ref{app:warm_starts}-\ref{app:special_cases} provide warm-start
and finite-step solver details, computational complexity, and the ablation mapping.}

\subsection{Theoretical Analysis}
\label{sec:theory_main}

We characterize optimistic selection and local response shaping, and derive
a Pareto-stationarity bound on an original lower-solution branch. Proofs and assumptions
are deferred to Appendix~\ref{app:theory}. For fixed $(\bx,\balpha)$ and
$\rho>0$, let
$\widetilde g_\rho(\bx,\by;\balpha)
:=g(\bx,\by)+\rho\psi_{\balpha}(\bx,\by)$ and define
$
\by^{\mathrm{glob}}_{\rho,\balpha}(\bx)
\in\argmin_{\by\in\cY}\widetilde g_\rho(\bx,\by;\balpha).
$
\rev{Proposition~\ref{prop:optimistic_consistency} explains why the selected
objective implements optimistic refinement of the original response set
as $\rho$ vanishes.}
\rev{Proposition~\ref{prop:response_shift} explains why a small selection
perturbation can be useful: it improves the selected upper-level criterion
at first order while increasing the original training loss only at second
order.}

\begin{proposition}[Optimistic-selection consistency]
\label{prop:optimistic_consistency}
Under Assumption~\ref{assump:existence}, as $\rho\downarrow0$, every
accumulation point of $\by^{\mathrm{glob}}_{\rho,\balpha}$ belongs to
$\argmin_{\by\in\argmin_{\bz}g(\bx,\bz)}
\psi_{\balpha}(\bx,\by)$.
\end{proposition}

\begin{proposition}[First-order preference gain]
\label{prop:response_shift}
Let $\by^0$ be an interior strict local minimizer of $g(\bx,\cdot)$ with
$G:=\nabla_{\by\by}^2g(\bx,\by^0)\succ0$. If $g$ and $f_s$ are three-times
continuously differentiable near $(\bx,\by^0)$, then, for sufficiently small
$\rho$, there exists a differentiable selected-response branch satisfying
{\small
\begin{equation}
\label{eq:response_shift}
    \by_{\rho,\balpha}
    =
    \by^0
    -
    \rho G^{-1}\ba_{\balpha}
    +
    \mathcal O(\rho^2),
    \qquad
    \ba_{\balpha}
    :=
    \nabla_{\by}\psi_{\balpha}(\bx,\by^0).
\end{equation}
\begin{equation}
    \psi_{\balpha}(\bx,\by_{\rho,\balpha})
    -
    \psi_{\balpha}(\bx,\by^0)
    =
    -\rho
    \ba_{\balpha}^\top G^{-1}\ba_{\balpha}
    +\mathcal O(\rho^2),~
    g(\bx,\by_{\rho,\balpha})
    -
    g(\bx,\by^0)
    =
    \frac{\rho^2}{2}
    \ba_{\balpha}^\top G^{-1}\ba_{\balpha}
    +\mathcal O(\rho^3). 
\end{equation}
}
\end{proposition}

For $\ba_{\balpha}\ne0$, the local perturbation improves the selected
criterion at first order, changes $g$ at second order, and displaces
the response by $\mathcal O(\rho)$.


Let $\bH_0(\bx)$ collect the upper-level hypergradients along a smooth
branch $\by^0(\bx)$ of the original lower problem, and define
$s(H):=\min_{\bw\in\simplex}\|H\bw\|$. This residual is zero at
Pareto-stationary points of the upper objectives on the branch.

\begin{theorem}[Pareto stationarity on an original solution branch]
\label{thm:variation_stationarity}
Consider $N$ PAOS iterations after warm-up with fixed, sufficiently small
$\rho>0$ and the step size and coordinator conditions of
Assumption~\ref{assump:pareto_descent}. Suppose the exact selected responses
follow a regular local continuation of a smooth branch $\by^0(\bx)$,
with $\by^0(\bx)\in\cY^\star(\bx)$ in a neighborhood of the upper iterates.
Suppose the hypergradient perturbation bound in
Corollary~\ref{cor:geometry_shift} holds along the iterates with a uniform
constant $C_H$.
Let $r_{\min}:=\min_s r_s>0$. Then
\begin{equation}
\label{eq:original_branch_average_bound}
    \frac{1}{N}\sum_{t=0}^{N-1}\E\!\left[s(\bH_0(\bx_t))^2\right]
    \leq \frac{2}{r_{\min}^2}\mathcal E_N+2C_H^2\rho^2,
\end{equation}
where $\mathcal E_N$ is the finite-iteration bound in
Lemma~\ref{lem:selected_direction_bound}, which accounts for the initial merit
gap, selector drift, direction bias, and stochastic noise.
\end{theorem}

At an exact fixed point of the selected dynamics, the pointwise residual
on the original branch is at most $C_H\rho$
(Eq.~\eqref{eq:original_stationarity_link}).
Appendix~\ref{app:stationarity_proof} derives the finite-iteration result.


\def\PAOSIncludeMain{1}


\ifdefined\PAOSIncludeMain

\section{Experiments}
\label{sec:experiments}

\paragraph{Benchmarks.}
We evaluate PAOS on an analytical double-well problem and four multi-objective
bilevel learning benchmarks. \emph{Data hyper-cleaning} uses MNIST to learn a
shared sample-weighting policy across five label-corruption levels.
\emph{Few-shot bilevel learning} uses a fixed 5-way 5-shot
support-query task from FC-100~\citep{oreshkin2018tadam}, coordinating five
class-wise losses on the query set used for upper-level optimization.
\emph{Physics-informed learning} reweights residual, initial-condition,
boundary, and observation points for a neural solver of the one-dimensional heat
equation~\citep{raissi2019physics}. \emph{Medical segmentation} learns the
training-loss composition of a compact U-Net on Kvasir-SEG~\citep{jha2020kvasir},
balancing regional overlap, boundary fidelity, and small-lesion quality. These
tasks cover sample weighting, representation learning, physical training-point
reweighting, and loss composition as upper-level decisions.
\rev{Table~\ref{tab:benchmark_bilevel_mapping} summarizes the upper and lower
variables and objectives for each learning task.}
Appendix~\ref{app:experimental_details} provides task formulations, data
splits, models, metrics, and implementation details.

\begin{table}[!ht]
    \centering
    \caption{\textcolor{RevisionBlue}{Multi-objective bilevel formulations of four learning benchmarks (Eq.~\eqref{eq:mobl}). Hyper-training selects $\bx$ to balance the upper objectives $\bF=(f_s)_{s=1}^{S}$, evaluated on the lower response $\by$ formed by training under $g$.}}
    \label{tab:benchmark_bilevel_mapping}
    \scriptsize
    \setlength{\tabcolsep}{1.5pt}
    \renewcommand{\arraystretch}{1.0}
    {\color{RevisionBlue}
    \begin{tabular}{|>{\bfseries\raggedright\arraybackslash}m{0.21\textwidth}|>{\raggedright\arraybackslash}m{0.17\textwidth}|>{\raggedright\arraybackslash}m{0.16\textwidth}|>{\raggedright\arraybackslash}m{0.21\textwidth}|>{\raggedright\arraybackslash}m{0.20\textwidth}|}
        \hline
        Task & \textbf{Upper variable $\bx$} & \textbf{Lower variable $\by$}
        & \textbf{Training objective $g$} & \textbf{Upper objectives $\bF$} \\ \hline
        Data hyper-cleaning
        & Shared sample weights
        & Five classifiers
        & Weighted cross-entropy\newline\mbox{(App.~\ref{app:mnist_hypercleaning}, Eq.~\eqref{eq:mnist_lower})}
        & Five clean-validation losses\newline\mbox{(App.~\ref{app:mnist_hypercleaning}, Eq.~\eqref{eq:mnist_upper})} \\ \hline
        Few-shot bilevel learning
        & Shared encoder
        & Task classifier
        & Support-set loss\newline\mbox{(App.~\ref{app:fc100}, Eq.~\eqref{eq:fc100_lower})}
        & Five class-wise query losses\newline\mbox{(App.~\ref{app:fc100}, Eq.~\eqref{eq:fc100_upper})} \\ \hline
        Physics-informed learning
        & Training-point weights
        & Neural solution
        & Weighted physical losses\newline\mbox{(App.~\ref{app:pinn}, Eq.~\eqref{eq:pinn_lower})}
        & Physics/observation losses\newline\mbox{(App.~\ref{app:pinn}, Eq.~\eqref{eq:pinn_upper})} \\ \hline
        Medical segmentation
        & Training-loss weights
        & Segmentation network
        & Weighted segmentation loss\newline\mbox{(App.~\ref{app:kvasir_definition}, Eq.~\eqref{eq:kvasir_lower})}
        & Region, edge, small lesion\newline\mbox{(App.~\ref{app:kvasir_definition}, Eq.~\eqref{eq:kvasir_upper})} \\ \hline
    \end{tabular}
    }
\end{table}

\begin{figure*}[t]
	\centering
	\includegraphics[width=\textwidth]{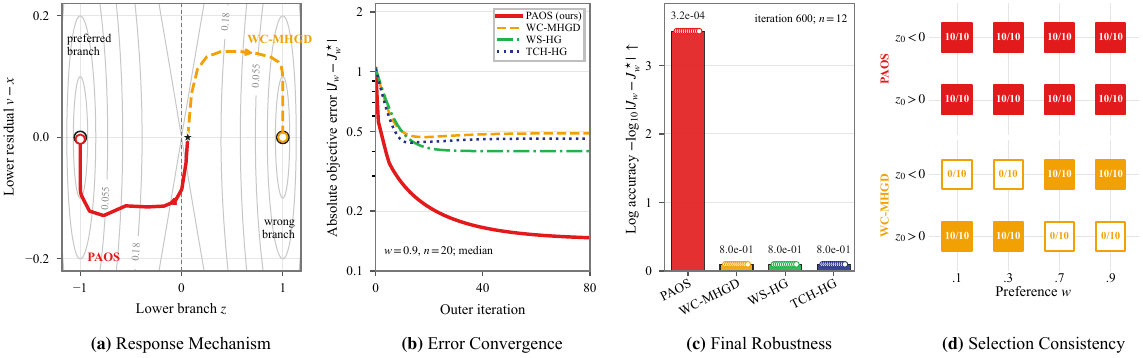}
	\vspace{-0.6cm}
\caption{\textbf{Pareto feedback shapes response attraction basins.}
	(a,b,d) Near-boundary branch selection on the 80-step problem;
	(b) uses weighted-sum reference regret. (c) Long-horizon refinement on a
	separate aligned-objective control (Appendix~\ref{app:dw_aligned}).}
	\label{fig:double_well_mechanism}
\end{figure*}

\begin{table*}[t]
	\centering
	\caption{\textbf{MNIST test accuracy.} Best and second-best values in each row are \textbf{bold} and
		\underline{underlined}.}
	\label{tab:mnist_hypercleaning}
	\scriptsize
	\setlength{\tabcolsep}{3pt}
	\renewcommand{\arraystretch}{1.08}
	\resizebox{\textwidth}{!}{%
		\begin{tabular}{|l|c|c|c|c|c|c|c|c|}
			\hline
			Corruption & MOML & FORUM & gMOBA & MGDA-HG & WS-HG & TCH-HG
			& WC-MHGD & \textbf{PAOS (ours)} \\ \hline
			$0\%$ $\uparrow$ & $81.36{\pm}0.08$ & $73.21{\pm}1.09$
			& $84.14{\pm}0.08$ & $84.04{\pm}0.00$ & $84.04{\pm}0.00$
			& $84.04{\pm}0.00$ & $\underline{86.98{\pm}0.03}$
			& $\mathbf{87.14{\pm}0.00}$ \\
			$10\%$ $\uparrow$ & $81.20{\pm}0.28$ & $73.20{\pm}0.28$
			& $83.69{\pm}0.38$ & $83.67{\pm}0.30$ & $83.59{\pm}0.01$
			& $83.59{\pm}0.01$ & $\underline{86.05{\pm}0.13}$
			& $\mathbf{86.20{\pm}0.03}$ \\
			$20\%$ $\uparrow$ & $80.99{\pm}0.47$ & $73.45{\pm}0.21$
			& $83.45{\pm}0.58$ & $83.35{\pm}0.44$ & $83.20{\pm}0.06$
			& $83.20{\pm}0.06$ & $\underline{84.98{\pm}0.06}$
			& $\mathbf{85.22{\pm}0.11}$ \\
			$30\%$ $\uparrow$ & $79.79{\pm}0.52$ & $73.25{\pm}0.16$
			& $82.29{\pm}0.18$ & $82.20{\pm}0.08$ & $82.56{\pm}0.14$
			& $82.56{\pm}0.14$ & $\underline{83.07{\pm}0.13}$
			& $\mathbf{83.40{\pm}0.11}$ \\
			$40\%$ $\uparrow$ & $79.73{\pm}0.10$ & $73.06{\pm}0.00$
			& $\mathbf{81.77{\pm}0.16}$ & $81.53{\pm}0.16$
			& $81.13{\pm}0.58$ & $81.13{\pm}0.58$ & $80.89{\pm}0.18$
			& $\underline{81.68{\pm}0.17}$ \\
			\hline
			Avg. $\uparrow$ & $80.61{\pm}0.03$ & $73.23{\pm}0.09$
			& $83.07{\pm}0.28$ & $82.96{\pm}0.20$ & $82.90{\pm}0.13$
			& $82.90{\pm}0.13$ & $\underline{84.39{\pm}0.02}$
			& $\mathbf{84.73{\pm}0.07}$ \\
			\hline
		\end{tabular}%
	}
\end{table*}

\begin{figure*}[t]
	\centering
	\includegraphics[width=\textwidth]{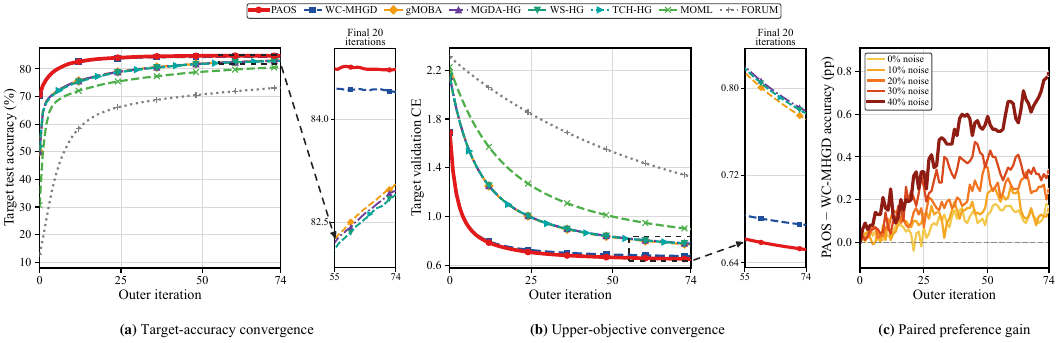}
	\vspace{-0.7cm}
	\caption{\textbf{MNIST hyper-cleaning trajectories.}
		\textbf{(a)} Target-aligned accuracy; \textbf{(b)} target validation
		cross-entropy; \textbf{(c)} paired accuracy gains over WC-MHGD across
		corruption preferences. }
	\label{fig:mnist_hypercleaning_curves}
\end{figure*}

\paragraph{Baselines and implementation.}
We compare PAOS with MOML~\citep{ye2021moml}, FORUM~\citep{ye2024forum},
gMOBA~\citep{yang2024gmoba}, MGDA-HG~\citep{sener2018multi,zhang2026mobl},
and WC-MHGD~\citep{zhang2026mobl}; weighted-sum and Chebyshev hypergradient
controls are denoted WS-HG and TCH-HG.
All experiments are implemented in PyTorch and run on NVIDIA RTX A6000 GPUs.

\paragraph{Mechanism analysis: Pareto-guided response formation.}
We isolate response selection on an analytical double-well problem with two
equally valid lower-level branches. The upper preference $w$ determines the
preferred branch:
\begin{equation} 
\mathcal S(x)=\{(-1,x),(+1,x)\},\qquad
z_w^\star=-\operatorname{sign}(w-\tfrac12),\qquad
z_{\mathrm{sep}}(\alpha)=\tfrac{3\rho}{4}(2\alpha_1-1). 
\label{eq:dw_mechanism}
\end{equation}
PAOS feeds the evolving Pareto mixture into $\alpha_t$, shifting the attraction
boundary before the response settles. Across the tested preferences and paired
near-boundary starts, PAOS consistently reaches the valid preferred branch.
At $w=0.9$, it also achieves the lowest median projected regret.
Figure~\ref{fig:double_well_mechanism} and Appendix~\ref{app:double_well}
further examine long-horizon refinement on an aligned-objective control
and the effect of initialization distance on branch selection.
These results show that Pareto feedback steers the formation of valid
lower-level responses toward the upper preference.

\paragraph{Data hyper-cleaning results.}
On MNIST, a shared sample-weighting policy coordinates five corruption-dependent
validation objectives.
Table~\ref{tab:mnist_hypercleaning} shows that PAOS achieves the highest
target-aligned test accuracy at four corruption levels and the best average
accuracy. Its gain over WC-MHGD is largest at $40\%$ corruption.
Figure~\ref{fig:mnist_hypercleaning_curves} shows higher target accuracy and
lower validation cross-entropy during optimization, together with paired
accuracy gains across corruption preferences.
Appendix~\ref{app:mnist_hypercleaning} details the task formulation and
evaluation protocol.

\begin{table*}[t]
    \centering
\caption{\textbf{Component ablation on MNIST hyper-cleaning.} Reports target-aligned test accuracy (\%) and target validation CE across 5 corruption levels. Best in \textbf{bold}, second in \underline{underlined}.}
    \label{tab:mnist_ablation}
    \small
    \setlength{\tabcolsep}{4.2pt}
    \resizebox{\textwidth}{!}{%
    \begin{tabular}{|l|c|c|c|c|c|c|c|}
        \hline
        Variant
        & $0\%$ $\uparrow$
        & $10\%$ $\uparrow$
        & $20\%$ $\uparrow$
        & $30\%$ $\uparrow$
        & $40\%$ $\uparrow$
        & Avg.\ Acc. $\uparrow$
        & Target Val.\ CE $\downarrow$ \\
        \hline
        Open loop ($\rho_t=0$)
        & $86.98{\pm}0.03$
        & $86.05{\pm}0.13$
        & $84.98{\pm}0.06$
        & $83.07{\pm}0.13$
        & $80.89{\pm}0.18$
        & $84.39{\pm}0.02$
        & $0.6746{\pm}0.0042$ \\

        Uniform selection ($\alpha_t=\mathbf{1}/S$)
        & $87.01{\pm}0.04$
        & $86.08{\pm}0.14$
        & $85.04{\pm}0.06$
        & $83.11{\pm}0.16$
        & $81.13{\pm}0.13$
        & $84.47{\pm}0.05$
        & $0.6699{\pm}0.0042$ \\

        Static preference ($\alpha_t=\mathbf r$)
        & $\underline{87.12{\pm}0.00}$
        & $\underline{86.17{\pm}0.01}$
        & $\underline{85.13{\pm}0.04}$
        & $\underline{83.32{\pm}0.14}$
        & $\underline{81.54{\pm}0.00}$
        & $\underline{84.66{\pm}0.04}$
        & $\underline{0.6565{\pm}0.0041}$ \\
        \textbf{Full PAOS}
        & $\mathbf{87.14{\pm}0.00}$
        & $\mathbf{86.20{\pm}0.03}$
        & $\mathbf{85.22{\pm}0.11}$
        & $\mathbf{83.40{\pm}0.11}$
        & $\mathbf{81.68{\pm}0.17}$
        & $\mathbf{84.73{\pm}0.07}$
        & $\mathbf{0.6522{\pm}0.0041}$ \\
        \hline
    \end{tabular}%
    }
\end{table*}

\begin{figure}[!t]
	\centering
	\includegraphics[width=\textwidth]{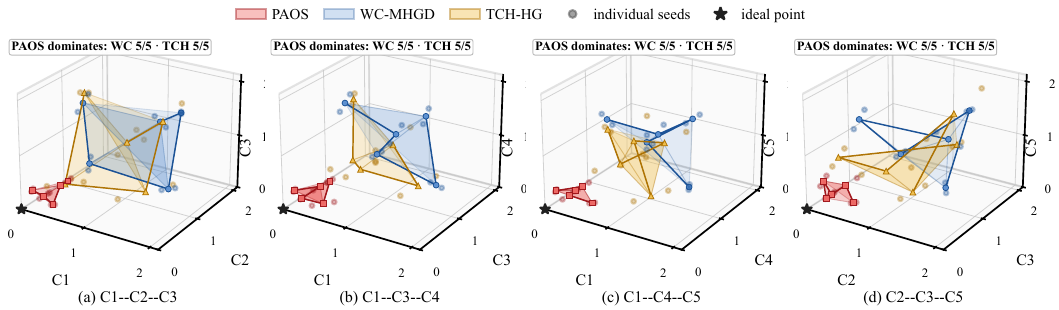}
	\vspace{-0.7cm}
	\caption{\textbf{FC-100 preference-conditioned trade-off projections.}
		Four three-objective views show five class-wise query preferences. The other six views appear
		in Appendix Figure~\ref{fig:fc100_pareto_appendix}. }
	\label{fig:fc100_pareto_main}
\end{figure}

\begin{table}[!t]
	\centering
	\caption{\textbf{FC-100: accuracy on the optimization query set.} Best in \textbf{bold}, second in \underline{underlined}.}
	\label{tab:fc100}
	\small
	\setlength{\tabcolsep}{3.5pt}
	\resizebox{\textwidth}{!}{%
		\begin{tabular}{|l|c|c|c|c|c|c|c|c|}
			\hline
			Metric & MOML & FORUM & gMOBA & MGDA-HG & WS-HG & TCH-HG
			& WC-MHGD & \textbf{PAOS (ours)} \\ \hline
			Final Acc. $\uparrow$
			& $64.67 \pm 12.26$ & $30.00 \pm 10.37$ & $86.67 \pm 3.77$
			& $92.00 \pm 3.77$ & $74.67 \pm 2.64$ & $74.00 \pm 7.35$
			& $\underline{95.47 \pm 0.00}$ & $\mathbf{96.27 \pm 1.13}$ \\
			Last-100 Acc. $\uparrow$
			& $64.92 \pm 12.61$ & $30.29 \pm 6.97$ & $84.23 \pm 4.64$
			& $92.67 \pm 0.20$ & $72.30 \pm 2.42$ & $67.35 \pm 4.14$
			& $\underline{94.30 \pm 0.37}$ & $\mathbf{94.85 \pm 0.67}$ \\
			Acc. AUC $\uparrow$
			& $64.43 \pm 12.38$ & $32.79 \pm 4.94$ & $70.53 \pm 1.81$
			& $78.09 \pm 1.58$ & $56.64 \pm 0.30$ & $51.55 \pm 1.20$
			& $\underline{80.30 \pm 0.60}$ & $\mathbf{80.52 \pm 0.80}$ \\
			\hline
		\end{tabular}%
	}
\end{table}

\begin{table}[!t]
    \centering
    \caption{\textbf{Heat-equation PINN point-reweighting: test relative $\bm{L}_2$ error.} Best and second-best values are
    \textbf{bold} and \underline{underlined}.}
    \label{tab:pinn_main}
    \small
    \setlength{\tabcolsep}{3.5pt}
    \resizebox{\textwidth}{!}{%
    \begin{tabular}{|l|c|c|c|c|c|c|c|c|}
        \hline
        Metric & MOML & FORUM & gMOBA & MGDA-HG & WS-HG & TCH-HG
        & WC-MHGD & \textbf{PAOS (ours)} \\ \hline
        Test rel. $L_2$ $\downarrow$
        & $0.055 \pm 0.008$ & $0.411 \pm 0.050$
        & $\underline{0.032 \pm 0.001}$ & $0.039 \pm 0.009$
        & $0.032 \pm 0.001$ & $0.032 \pm 0.000$
        & $0.055 \pm 0.034$ & $\mathbf{0.028 \pm 0.000}$ \\
        Rel.-$L_2$ AUC $\downarrow$
        & $0.128 \pm 0.000$ & $0.484 \pm 0.060$
        & $0.087 \pm 0.006$ & $0.087 \pm 0.006$
        & $0.087 \pm 0.007$ & $\underline{0.087 \pm 0.007}$
        & $0.088 \pm 0.008$ & $\mathbf{0.082 \pm 0.005}$ \\
        Final-20 rel. $L_2$ $\downarrow$
        & $0.062 \pm 0.009$ & $0.423 \pm 0.051$
        & $0.051 \pm 0.021$ & $0.052 \pm 0.022$
        & $\underline{0.051 \pm 0.022}$ & $0.051 \pm 0.022$
        & $0.054 \pm 0.026$ & $\mathbf{0.040 \pm 0.003}$ \\
        \hline
    \end{tabular}%
    }
\end{table}

\paragraph{Ablation studies.}
Table~\ref{tab:mnist_ablation} separates the contributions of response
selection, static preference guidance, and dynamic Pareto feedback.
Uniform selection improves on the open-loop baseline, and static preference
guidance yields a larger gain. Feeding back the evolving Pareto mixture
further improves mean accuracy at every corruption level, giving full PAOS
the highest average accuracy and lowest target validation CE.
Across the ten preference--seed pairs, full PAOS improves accuracy in nine
and ties in one, while reducing target validation CE in all ten.
These results support the complementary roles of preference-guided response
selection and dynamic feedback. Feedback-rate sweeps also show improved
valid branch selection in the double well and lower PINN prediction error
with dynamic feedback than with static selection
(Appendix Tables~\ref{tab:dw_beta_ablation}
and~\ref{tab:pinn_beta_ablation}).

\paragraph{Few-shot bilevel learning results.}
On the fixed 5-way 5-shot task, PAOS shapes finite-budget classifier responses
to improve the five objectives on the optimization query set. PAOS dominates
all five preference means of both WC-MHGD and TCH-HG in each of the ten
three-objective projections (Figure~\ref{fig:fc100_pareto_main} and Appendix
Figure~\ref{fig:fc100_pareto_appendix}). This trade-off advantage is
complemented by the best final, last-100, and trajectory-average query
accuracies (Table~\ref{tab:fc100}), combining improved trade-offs with
stronger optimization performance.

\paragraph{Physics-informed learning results.}
We study point reweighting for a heat-equation PINN with residual,
initial-condition, boundary-condition, and observation objectives. Across
terminal, trajectory-average, and late-stage relative $L_2$ errors,
PAOS achieves the lowest errors (Table~\ref{tab:pinn_main}). Holding all other
settings fixed, dynamic feedback at $\beta=0.25$ reduces relative $L_2$ error
by $23.6\%$ and test MSE by $49.4\%$ over static selection ($\beta=0$), with
paired improvements in $6/8$ test runs and $7/8$ validation runs.
These results show the benefit of adapting response selection as
response-conditioned trade-offs evolve.
Appendix~\ref{app:pinn} reports the complete feedback-rate sweep.

\begin{table}[t]
    \centering
    \caption{\textbf{Kvasir-SEG uniform-preference results with a compact U-Net.} Best and second-best values in each column are \textbf{bold} and \underline{underlined}.}
    \label{tab:kvasir_main}
    \small
    \setlength{\tabcolsep}{3.2pt}
    \resizebox{\linewidth}{!}{%
    \begin{tabular}{|l|c|c|c|c|c|}
        \hline
        Method & Dice $\uparrow$ & IoU $\uparrow$ & Boundary F1 $\uparrow$
               & Small Dice $\uparrow$ & HD95 $\downarrow$ \\ \hline
        MOML    & $0.4281 \pm 0.0560$ & $0.2996 \pm 0.0493$ & $0.3294 \pm 0.0257$
                & $0.2374 \pm 0.0415$ & $32.6203 \pm 2.6727$ \\
        FORUM   & $0.3612 \pm 0.0168$ & $0.2416 \pm 0.0150$ & $0.2386 \pm 0.0283$
                & $0.1305 \pm 0.0033$ & $46.4731 \pm 4.2473$ \\
        WC-MHGD & $\underline{0.4711 \pm 0.0314}$ & $\underline{0.3422 \pm 0.0266}$ & $\mathbf{0.3676 \pm 0.0135}$
                & $\underline{0.2496 \pm 0.0199}$ & $\underline{32.4722 \pm 1.2692}$ \\
        PAOS    & $\mathbf{0.4849 \pm 0.0231}$ & $\mathbf{0.3535 \pm 0.0189}$
                & $\underline{0.3611 \pm 0.0239}$ & $\mathbf{0.2623 \pm 0.0278}$
                & $\mathbf{31.6363 \pm 1.2706}$ \\
        \hline
    \end{tabular}}
\end{table}

\paragraph{Medical segmentation results.}
On Kvasir-SEG, PAOS achieves the highest Dice, IoU, and small-lesion Dice,
together with the lowest HD95 under the uniform preference
(Table~\ref{tab:kvasir_main}). These gains reflect improved regional
overlap, small-lesion recovery, and boundary distance over WC-MHGD.
Figure~\ref{fig:kvasir_qualitative} illustrates the resulting masks;
Appendix~\ref{app:kvasir_additional} provides further lesion-size comparisons.

\begin{figure}[!t]
    \centering
    \includegraphics[width=\linewidth]{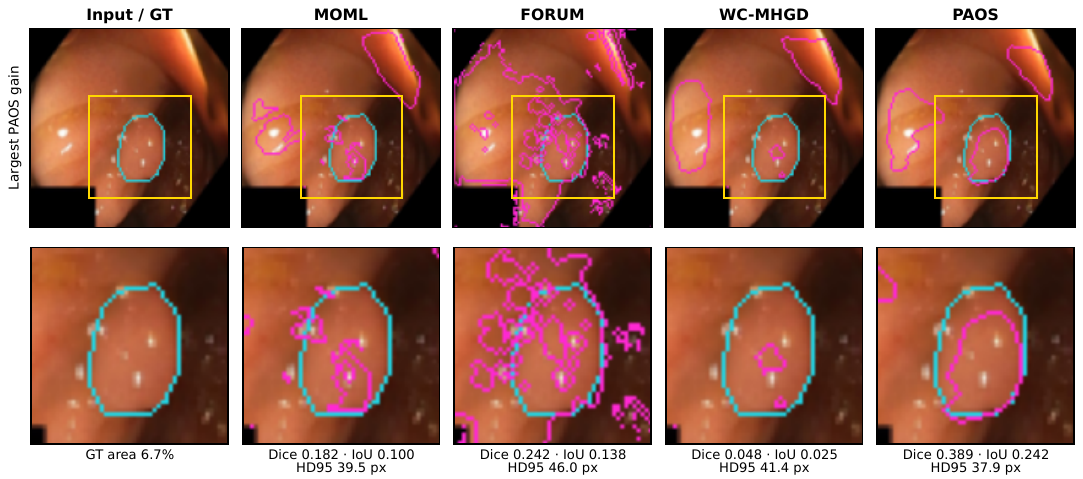}
     \vspace{-0.7cm}
    \caption{Qualitative Kvasir-SEG comparison on the mid-size test case with
    the largest PAOS gain.}
    \label{fig:kvasir_qualitative}
\end{figure}

\paragraph{Quality-efficiency analysis.}
Table~\ref{tab:quality_efficiency} shows that PAOS achieves the highest
target accuracy and the lowest final CE and CE AUC.
It reaches CE $\leq0.8$ in 11 iterations, taking 75.2 seconds.
At this threshold, PAOS is approximately $2.8\times$ faster in wall-clock
time than gMOBA and MGDA-HG, while using less than one-tenth of the
allocated memory of unrolled MOML.

\fi


\ifdefined\PAOSIncludeAppendix

\section{Experimental Details}
\label{app:experimental_details}

This appendix details the bilevel formulations, datasets, models,
optimization settings, and evaluation metrics for
Section~\ref{sec:experiments}.
For each learning benchmark, $g(x,y)$ is the scalar lower-level
training objective, and
$\mathbf F(x,y)=(f_1(x,y),\ldots,f_S(x,y))^\top$
collects the upper-level objectives. They induce
$\bPhi(x)=\mathbf F(x,y^\star(x))$, where
$y^\star(x)\in\argmin_y g(x,y)$.
PAOS forms finite-step responses using the selected objective in
Eq.~\eqref{eq:selected_lower}.

\subsection{Preferences and Implementation}
\label{app:common_protocol}

\paragraph{Baseline implementation.}
We implement MOML, FORUM, gMOBA, MGDA-HG, WS-HG, TCH-HG, WC-MHGD,
and PAOS according to their published update rules.
Learning rates are selected separately for each method and task;
the accompanying code provides the complete configurations.
For gMOBA, we use the single-lower-objective specialization.

\paragraph{Computing environment.}
Learning experiments run on Linux with two NVIDIA RTX A6000 GPUs, Python 3.10.20,
PyTorch 2.5.1, CUDA 12.1, and cuDNN 9.1.

\subsection{MNIST Data Hyper-Cleaning}
\label{app:mnist_hypercleaning}

\paragraph{Task construction.}
We partition MNIST into 5,000 lower-level training images, 5,000 clean
validation images, and 5,000 test images. Each image is flattened into a
784-dimensional vector. Five training environments use label corruption
rates $c_s\in\{0,0.1,0.2,0.3,0.4\}$, $s\in\{1,\ldots,5\}$.
Let
$\mathcal{D}_{\mathrm{tr}}^{(s)}
=\{(a_i,\widetilde b_i^{(s)})\}_{i=1}^{n}$
denote the training set whose labels are corrupted at rate $c_s$.

The upper variable $x\in\mathbb{R}^{n}$ contains one cleaning logit for each
training example, and $\sigma(x_i)$ determines its contribution to lower-level
training. The lower response $y=(y_1,\ldots,y_5)$ consists of five linear
softmax classifiers, one for each corruption environment. The original
lower-level objective is
\begin{equation}
    g(x,y)
    ={}\frac{1}{5}\sum_{s=1}^{5}
        \frac{1}{n}\sum_{i=1}^{n}\sigma(x_i)
        \ell_{\mathrm{ce}}
        \bigl(h_{y_s}(a_i),\widetilde b_i^{(s)}\bigr)
    +\frac{\lambda_y}{10}\sum_{s=1}^{5}\|y_s\|_2^2
    +\frac{\lambda_x}{2}\|x\|_2^2 .
    \label{eq:mnist_lower}
\end{equation}
Each upper-level objective evaluates one corruption-specific classifier on the
shared clean validation set:
\begin{equation}
    f_s(x,y)
    =\frac{1}{|\mathcal{D}_{\mathrm{val}}|}
    \sum_{(a,b)\in\mathcal{D}_{\mathrm{val}}}
    \ell_{\mathrm{ce}}\bigl(h_{y_s}(a),b\bigr),
    \hspace{1mm}s\in\{1,\ldots,5\}.
\label{eq:mnist_upper}
\end{equation}
These five losses form
$\mathbf F(x,y)=(f_1(x,y),\ldots,f_5(x,y))^\top$;
the shared sample weights affect them through the trained classifiers.
For fixed sample weights, positive quadratic regularization makes the
linear-classifier lower problem strongly convex with a unique minimizer.
PAOS shapes its finite-budget responses through the selected training
objective.

\paragraph{Preference-conditioned evaluation.}
For target corruption environment $s$, we use
\begin{equation}
    r_j^{(s)}
    =
    \begin{cases}
        0.8, & j=s,\\
        0.05, & j\neq s,
    \end{cases}
    \qquad
    \sum_{j=1}^{5}r_j^{(s)}=1.
    \label{eq:mnist_preferences}
\end{equation}
For a preference-conditioned method, target-aligned accuracy at
corruption level $c_s$ is the test accuracy of $h_{y_s}$ from the run
with $r^{(s)}$.
For a preference-independent method, it is the accuracy of the
corresponding classifier from the shared multi-objective run.
Table~\ref{tab:mnist_hypercleaning} reports the average
\begin{equation}
    \operatorname{Acc}_{\mathrm{avg}}
    =\frac{1}{5}\sum_{s=1}^{5}
    \operatorname{Acc}_{\mathrm{test}}
    \bigl(h_{y_s}^{(r^{(s)})}\bigr).
    \label{eq:mnist_target_accuracy}
\end{equation}
\paragraph{Optimization details.}
PAOS uses 75 outer iterations and 200 lower-level gradient steps per outer
iteration, with lower and upper learning rates of $0.05$ and $0.1$,
respectively, ten conjugate-gradient iterations, and damping $10^{-4}$.
Classifier weight decay is $10^{-3}$ and upper-logit regularization is
$10^{-5}$. For PAOS, the response-selection strength is linearly increased
from zero to $0.005$ during the first 25 outer iterations, and the
selection-state momentum is $0.5$.

\subsection{Few-shot Bilevel Learning on FC-100}
\label{app:fc100}

\paragraph{Dataset and task construction.}
FC-100~\citep{oreshkin2018tadam} partitions the CIFAR-100 classes into
disjoint training, validation, and test subsets.
We sample a 5-way 5-shot support--query task from the training classes,
using five support images and 15 query images per class.
Images are resized to $32\times32$ pixels and processed with a shared
augmentation and normalization pipeline.
Each run performs 500 outer iterations on one sampled task.
The query set supplies both the upper-level optimization losses and
the reported accuracies, measuring within-task optimization performance.

\paragraph{Bilevel formulation.}
Let $\phi_x$ denote a shared feature encoder parameterized by the upper-level
variable $x$, and let $h_y$ denote a task-specific linear classifier
parameterized by the lower-level variable $y$. Given a task with support
set $\mathcal{D}_{\mathrm{sup}}$, the lower-level classifier is adapted by
minimizing
\begin{equation}
    g(x,y)
    =\frac{1}{|\mathcal{D}_{\mathrm{sup}}|}
    \sum_{(a,b)\in\mathcal{D}_{\mathrm{sup}}}
    \ell_{\mathrm{ce}}(h_y(\phi_x(a)),b)
    +\frac{\lambda_y}{2}\|y\|_2^2 ,
    \label{eq:fc100_lower}
\end{equation}
Here $\ell_{\mathrm{ce}}$ is the cross-entropy loss and $\lambda_y$
controls classifier regularization.
With a fixed encoder and positive quadratic regularization, the lower
problem has a unique minimizer.
PAOS shapes the classifier response under the finite adaptation budget.
The upper level contains one query objective for each of the five classes:
\begin{equation}
    f_c(x,y)
    =\frac{1}{|\mathcal{D}^{c}_{\mathrm{qry}}|}
    \sum_{(a,b)\in\mathcal{D}^{c}_{\mathrm{qry}}}
    \ell_{\mathrm{ce}}(h_y(\phi_x(a)),b),
    \qquad c\in\{1,\ldots,5\}.
\label{eq:fc100_upper}
\end{equation}
Here $\mathcal{D}^{c}_{\mathrm{qry}}$ is the class-$c$ query subset.
Thus $\mathbf F(x,y)=(f_1(x,y),\ldots,f_5(x,y))^\top$.
The shared encoder $x$ enters both support-set training and the
class-wise query objectives.
The labeled query objectives enter both the upper-level update and
PAOS's selected lower objective.
For each task, we run five preferences, selecting each class once:
\begin{equation}
    r^{(c)}_j
    =
    \begin{cases}
        0.8,  & j=c,\\
        0.05, & j\neq c.
    \end{cases}
    \label{eq:fc100_preference}
\end{equation}
\paragraph{Model and optimization settings.}
The feature extractor is a four-layer convolutional network with 32 channels
per layer, followed by a task-specific linear classifier. In the PAOS
configuration, each
outer iteration performs 32 lower-level adaptation steps with learning rate
$0.2$, followed by an upper-level update with learning rate $0.01$.
Hypergradient computation uses five conjugate-gradient iterations with damping
coefficient $0.01$. We apply gradient-clipping thresholds of $10$ and $2$ to
the lower- and upper-level updates, respectively, and set the classifier weight
decay to $10^{-3}$.

For PAOS, the selection strength is linearly warmed from zero to
$2\times10^{-3}$ during the first 100 outer iterations. The response-selection
state is updated with momentum coefficient $0.5$.

\paragraph{Evaluation metrics.}
We report final, last-100, and trajectory-average query accuracies.
Final accuracy is measured after the last outer iteration.
Last-100 accuracy averages the final 100 iterations.
Accuracy AUC averages all 500 outer iterations:
\begin{equation}
    \operatorname{AUC}_{\mathrm{acc}}
    =\frac{1}{T}\sum_{t=1}^{T}
    \operatorname{Acc}_{\mathrm{qry}}^{(t)},
    \qquad T=500.
    \label{eq:fc100_auc}
\end{equation}
Table~\ref{tab:fc100} reports the mean and standard deviation across five
independent seeds.

\paragraph{Additional preference-conditioned trade-off projections.}
Figure~\ref{fig:fc100_pareto_appendix} completes the ten three-objective
projections with the six views omitted from Figure~\ref{fig:fc100_pareto_main}.

\begin{figure}[!t]
    \centering
    \includegraphics[width=\textwidth]{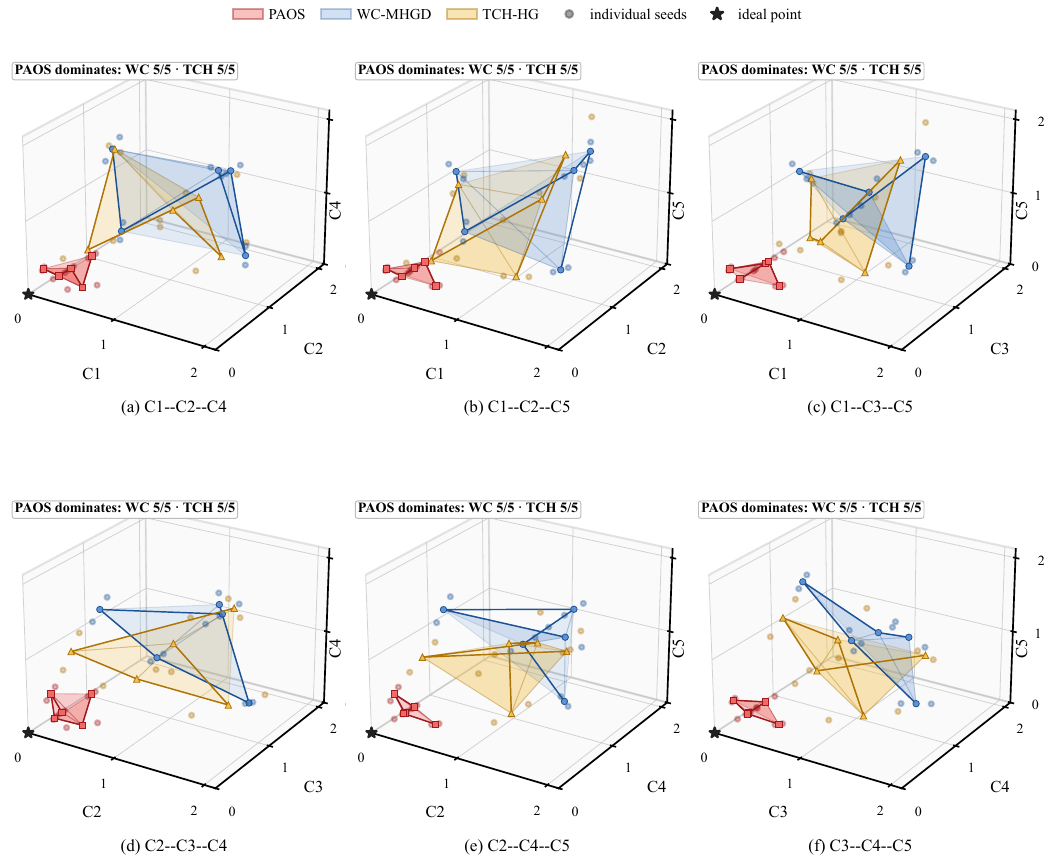}
    	\vspace{-0.6cm}
    \caption{\textbf{Additional FC-100 preference-conditioned trade-off
    projections.} These six views complete all ten three-objective projections.
    In every view, PAOS dominates all five preference means of both WC-MHGD and
    TCH-HG (lower is better); translucent markers show individual seeds.}
    \label{fig:fc100_pareto_appendix}
\end{figure}

\subsection{Physics-Informed Learning on the Heat Equation}
\label{app:pinn}

\paragraph{Problem construction.}
We consider the one-dimensional heat equation
\begin{equation}
    \frac{\partial u}{\partial t}
    =\nu\frac{\partial^2u}{\partial \xi^2},
    \qquad
    (\xi,t)\in[-1,1]\times[0,1],
    \qquad
    \nu=0.1,
    \label{eq:pinn_heat}
\end{equation}
with initial and boundary conditions
\begin{equation}
    u(\xi,0)=\sin(\pi\xi),
    \qquad
    u(-1,t)=u(1,t)=0.
\end{equation}
Its analytical solution is
\begin{equation}
    u^\star(\xi,t)
    =\exp(-\pi^2\nu t)\sin(\pi\xi).
    \label{eq:pinn_exact}
\end{equation}

The lower-level variable $y$ parameterizes a three-layer tanh multilayer
perceptron $u_y(\xi,t)$ with hidden width 32. Its physics residual is
\begin{equation}
    \mathcal R_y(\xi,t)
    =\frac{\partial u_y(\xi,t)}{\partial t}
    -\nu\frac{\partial^2u_y(\xi,t)}{\partial \xi^2}.
    \label{eq:pinn_residual}
\end{equation}

We construct four groups of pointwise training losses:
\begin{align}
    \ell_{\mathrm{res},i}(y)
    &=\left|\mathcal R_y(\xi_i,t_i)\right|^2,
    &\ell_{\mathrm{IC},i}(y)
    &=\left|u_y(\xi_i,0)-\sin(\pi\xi_i)\right|^2,\\
    \ell_{\mathrm{BC},i}(y)
    &=\left|u_y(\xi_i^{\mathrm{bd}},t_i)\right|^2,
    &\ell_{\mathrm{obs},i}(y)
    &=\left|u_y(\xi_i,t_i)-\widetilde u_i\right|^2,
\end{align}
Here $\xi_i^{\mathrm{bd}}\in\{-1,1\}$; noisy observations satisfy
\begin{equation}
    \widetilde u_i
    =u^\star(\xi_i,t_i)+\epsilon_i,
    \qquad
    \epsilon_i\sim\mathcal N(0,0.01^2).
\end{equation}

\paragraph{Point-reweighting bilevel formulation.}
The upper-level variable
$x=\{x_{k,i}:k=1,\ldots,4,\ i=1,\ldots,n_k\}$ contains one logit for every
training point. For constraint group $k$, we convert these logits into positive
mean-one weights:
\begin{equation}
    w_{k,i}(x)
    =n_k\frac{\exp(x_{k,i})}
    {\sum_{j=1}^{n_k}\exp(x_{k,j})},
    \qquad
    \frac{1}{n_k}\sum_{i=1}^{n_k}w_{k,i}(x)=1.
    \label{eq:pinn_point_weights}
\end{equation}
The resulting lower-level objective is
\begin{equation}
    g(x,y)
    =\sum_{k=1}^{4}\frac{1}{n_k}
    \sum_{i=1}^{n_k}w_{k,i}(x)\ell_{k,i}(y)
    +\frac{\lambda_y}{2}\|y\|_2^2
    +\frac{\lambda_x}{2}\|x\|_2^2.
    \label{eq:pinn_lower}
\end{equation}

The four upper-level objectives evaluate the corresponding constraints on
independently sampled, noise-free validation points:
\begin{equation}
    \mathbf F(x,y)
    =
    \begin{bmatrix}
        F_{\mathrm{res}}(x,y) \\
        F_{\mathrm{IC}}(x,y) \\
        F_{\mathrm{BC}}(x,y) \\
        F_{\mathrm{obs}}(x,y)
    \end{bmatrix}
    =
    \begin{bmatrix}
        \operatorname{MSE}^{\mathrm{val}}_{\mathrm{res}}(y) \\
        \operatorname{MSE}^{\mathrm{val}}_{\mathrm{IC}}(y) \\
        \operatorname{MSE}^{\mathrm{val}}_{\mathrm{BC}}(y) \\
        \operatorname{MSE}^{\mathrm{val}}_{\mathrm{obs}}(y)
    \end{bmatrix}.
    \label{eq:pinn_upper}
\end{equation}
For $k\in\{\mathrm{res},\mathrm{IC},\mathrm{BC},\mathrm{obs}\}$,
$\operatorname{MSE}^{\mathrm{val}}_k(y)
=|\mathcal V_k|^{-1}\sum_{i\in\mathcal V_k}
\ell^{\mathrm{val}}_{k,i}(y)$, where $\mathcal V_k$
indexes independently sampled validation points.
The observation-validation targets use $u^\star$ without training noise.
\paragraph{Data and preference protocol.}
The training set contains 256 PDE-residual points, 96 initial-condition points,
96 boundary-condition points, and 96 observation points, yielding 544
point-weighting variables in total.
Testing uses a dense $48\times96$ space--time grid.

We evaluate four preference directions, with each upper objective selected once
as the preferred objective. For preference $p\in\{1,\ldots,4\}$, the
simplex-normalized preference is
\begin{equation}
    r_k^{(p)}
    =
    \begin{cases}
        16/19, & k=p,\\
        1/19,  & k\neq p.
    \end{cases}
    \label{eq:pinn_preference}
\end{equation}
Results are reported over two independent seeds per preference direction.

\paragraph{Optimization settings.}
PAOS uses 80 outer iterations and 60 lower-level steps per outer iteration.
The lower- and upper-level learning rates are $0.01$ and $0.02$;
hypergradient estimation uses eight conjugate-gradient steps with damping
coefficient $10^{-3}$. The lower gradient and upper update direction are
clipped at norms 10 and 3, respectively.

For PAOS, the selection strength is linearly warmed from zero to $0.005$ during
the first 50 outer iterations, and the feedback rate is set to
$\beta=0.25$. We select $\beta$ from $\{0,0.25,0.5,0.7,1\}$ using the target-aligned
validation objective and report test metrics for the selected setting.

\paragraph{Evaluation metrics.}
Let $\mathcal G_{\mathrm{test}}$ denote the dense space--time test grid.
We report the relative $L_2$ error and its iteration-normalized AUC:
\begin{equation}
    E_{\mathrm{rel}}
    =\frac{\left\|u_y-u^\star\right\|_{2,\mathcal G_{\mathrm{test}}}}
    {\left\|u^\star\right\|_{2,\mathcal G_{\mathrm{test}}}},
    \qquad
    E_{\mathrm{AUC}}=\frac{1}{T}\sum_{s=1}^{T}E_{\mathrm{rel}}^{(s)},
    \quad T=80.
    \label{eq:pinn_rel_l2}
\end{equation}
We also report the average relative $L_2$ error over the final 20 iterations.

\paragraph{Feedback-rate ablation.}
Table~\ref{tab:pinn_beta_ablation} varies only $\beta$.
Compared with static selection, $\beta=0.25$ reduces relative $L_2$
error by $23.6\%$, test MSE by $49.4\%$, and the selected validation
objective by $22.9\%$, with improvements in $6/8$ paired test runs
and $7/8$ validation runs.
Moderate feedback achieves better prediction accuracy than immediate
target replacement.
Science score is $1/(1+E_{\mathrm{rel}})$, and selected validation
denotes the preference-indexed upper objective.

\begin{table*}[t]
    \centering
    \caption{PINN feedback-rate ablation over four preference directions and
    two seeds. Metrics are macro-averaged over preferences within each seed;
    only $\beta$ varies.}
    \label{tab:pinn_beta_ablation}
    \small
    \setlength{\tabcolsep}{5pt}
    \resizebox{\textwidth}{!}{%
    \begin{tabular}{|l|c|c|c|c|}
        \hline
        Feedback rate & Science score $\uparrow$ & Relative $L_2$ $\downarrow$
        & Test MSE $\times10^{-4}$ $\downarrow$
        & Selected val. $\times10^{-4}$ $\downarrow$ \\ \hline
        $\beta=0$ (static) & $0.96480\pm0.00105$ & $0.03674\pm0.00142$
        & $3.52\pm0.91$ & $8.46\pm2.74$ \\
        $\beta=0.25$ (PAOS) & $\mathbf{0.97274\pm0.00009}$
        & $\mathbf{0.02805\pm0.00012}$ & $\mathbf{1.78\pm0.08}$
        & $\mathbf{6.52\pm0.98}$ \\
        $\beta=0.50$ & $0.96570\pm0.00266$ & $0.03558\pm0.00290$
        & $2.90\pm0.55$ & $11.01\pm0.53$ \\
        $\beta=0.70$ & $0.97087\pm0.00456$ & $0.03005\pm0.00488$
        & $2.07\pm0.72$ & $7.76\pm1.66$ \\
        $\beta=1.00$ & $0.94370\pm0.03954$ & $0.06395\pm0.04912$
        & $20.10\pm25.66$ & $30.38\pm32.76$ \\
        \hline
    \end{tabular}
    }
\end{table*}

\FloatBarrier

\subsection{Kvasir-SEG Uniform-Preference Loss Design}
\label{app:kvasir_definition}

Kvasir-SEG contains 1,000 endoscopic images and corresponding polyp masks. We
stratify by lesion area and use a fixed $700/150/150$ train/validation/test split
with split seed 2027. Images and masks are resized to $96\times96$ using bilinear
and nearest-neighbor interpolation, respectively. Training augmentation consists
of random horizontal flips and lower-probability vertical flips.

The upper variable $x\in\mathbb R^3$ contains training-loss logits, with
$\omega(x)=3\operatorname{softmax}(x)$. The lower response $y$ comprises the
7,697 parameters of a compact U-Net with base width four. The lower objective is
\begin{equation}
    g(x,y)=\omega_1(x)\mathcal L_{\mathrm{BCE}}(y)
    +\omega_2(x)\mathcal L_{\mathrm{region}}(y)
    +\omega_3(x)\mathcal L_{\mathrm{boundary}}(y)
    +\frac{\lambda_y}{2}\|y\|^2,
    \label{eq:kvasir_lower}
\end{equation}
Here $\mathcal L_{\mathrm{region}}$ and
$\mathcal L_{\mathrm{boundary}}$ are the soft regional-Dice and
boundary-Dice losses on training minibatches, respectively.
Let $\mathcal V$ be a validation minibatch and
$\mathcal V_{\mathrm{small}}$ a minibatch drawn from the lowest
lesion-area quartile of the validation split. The upper-level
vector is
$\mathbf F(x,y)=(f_1(x,y),f_2(x,y),f_3(x,y))^\top$, with
\begin{equation}
\begin{aligned}
f_1(x,y)&=1-\operatorname{Dice}_{\mathrm{region}}(y;\mathcal V),\\
f_2(x,y)&=1-\operatorname{Dice}_{\mathrm{boundary}}(y;\mathcal V),\\
f_3(x,y)&=1-\operatorname{Dice}_{\mathrm{region}}
          (y;\mathcal V_{\mathrm{small}}).
\end{aligned}
\label{eq:kvasir_upper}
\end{equation}
All methods use the same uniform user preference.

PAOS uses 180 outer iterations, six lower steps, two CG steps, damping 0.05,
and lower and upper learning rates of 0.03. The selection strength is warmed to
$0.1$ over 20 iterations with selector momentum $\beta=0.7$. Training and
validation batch sizes are six, and the test batch size is 16.

Test metrics are Dice, IoU, Boundary F1 with a two-pixel tolerance, Dice on the
lowest lesion-area quartile, and the 95th-percentile Hausdorff distance (HD95).
HD95 is measured in resized-image pixels; metric directions appear in
Table~\ref{tab:kvasir_main}.

\section{Additional Experimental Results}
\label{app:additional_results}

\subsection{Quality-to-Target Efficiency}
\label{app:quality_efficiency}

\begin{table}[H]
    \centering
    \caption{MNIST hyper-cleaning quality-to-target efficiency over 75 outer
    iterations (mean $\pm$ std. over three seeds, RTX A6000). ``--'' means
    target CE $\leq 0.8$ was not reached. GPU memory is allocated memory in MiB.}
    \label{tab:quality_efficiency}
    \scriptsize
    \setlength{\tabcolsep}{2.6pt}
    \renewcommand{\arraystretch}{1.08}
    \resizebox{\textwidth}{!}{%
    \begin{tabular}{|l|c|c|c|c|c|c|c|c|}
        \hline
        & \multicolumn{3}{c|}{Solution quality}
        & \multicolumn{2}{c|}{Reach CE $\leq 0.8$}
        & GPU memory
        & \multicolumn{2}{c|}{Oracle calls / outer} \\ \cline{2-9}
        Method
        & Target Acc. $\uparrow$
        & Final CE $\downarrow$
        & CE AUC $\downarrow$
        & Iter. $\downarrow$
        & Time (s) $\downarrow$
        & Alloc. $\downarrow$
        & LL grad.
        & HVP \\ \hline
        MOML
        & $80.62{\pm}0.23$
        & $0.8752{\pm}0.0078$
        & $1.2099{\pm}0.0060$
        & --
        & --
        & $3133.4{\pm}0.4$
        & 20
        & 0 \\

        FORUM
        & $72.89{\pm}0.08$
        & $1.3322{\pm}0.0011$
        & $1.7275{\pm}0.0015$
        & --
        & --
        & $188.0{\pm}0.4$
        & 64
        & 0 \\

        gMOBA
        & $83.39{\pm}0.08$
        & $0.7490{\pm}0.0086$
        & $0.9932{\pm}0.0072$
        & $53.7{\pm}2.3$
        & $207.5{\pm}9.8$
        & $162.3{\pm}0.5$
        & 200
        & 50 \\

        MGDA-HG
        & $83.24{\pm}0.08$
        & $0.7561{\pm}0.0087$
        & $0.9969{\pm}0.0072$
        & $55.3{\pm}2.9$
        & $213.7{\pm}10.8$
        & $162.1{\pm}0.4$
        & 200
        & 50 \\

        WC-MHGD
        & $83.61{\pm}0.12$
        & $0.6923{\pm}0.0094$
        & $0.7846{\pm}0.0089$
        & $11.7{\pm}0.6$
        & $\mathbf{44.9{\pm}2.0}$
        & $162.3{\pm}0.4$
        & 200
        & 50 \\
        \textbf{PAOS}
        & $\mathbf{85.17{\pm}0.11}$
        & $\mathbf{0.6417{\pm}0.0089}$
        & $\mathbf{0.7200{\pm}0.0085}$
        & $\mathbf{11.0{\pm}0.0}$
        & $75.2{\pm}0.2$
        & $281.3{\pm}0.4$
        & 200
        & 50 \\
        \hline
    \end{tabular}%
    }
\end{table}

\subsection{Additional Kvasir-SEG Results}
\label{app:kvasir_additional}

Figure~\ref{fig:kvasir_quartiles} compares test cases across the four
ground-truth lesion-area quartiles.
Each row includes the full image, a shared region-of-interest crop,
and per-case Dice, IoU, and HD95.

\begin{figure}[!t]
	\centering
	\includegraphics[width=\textwidth]{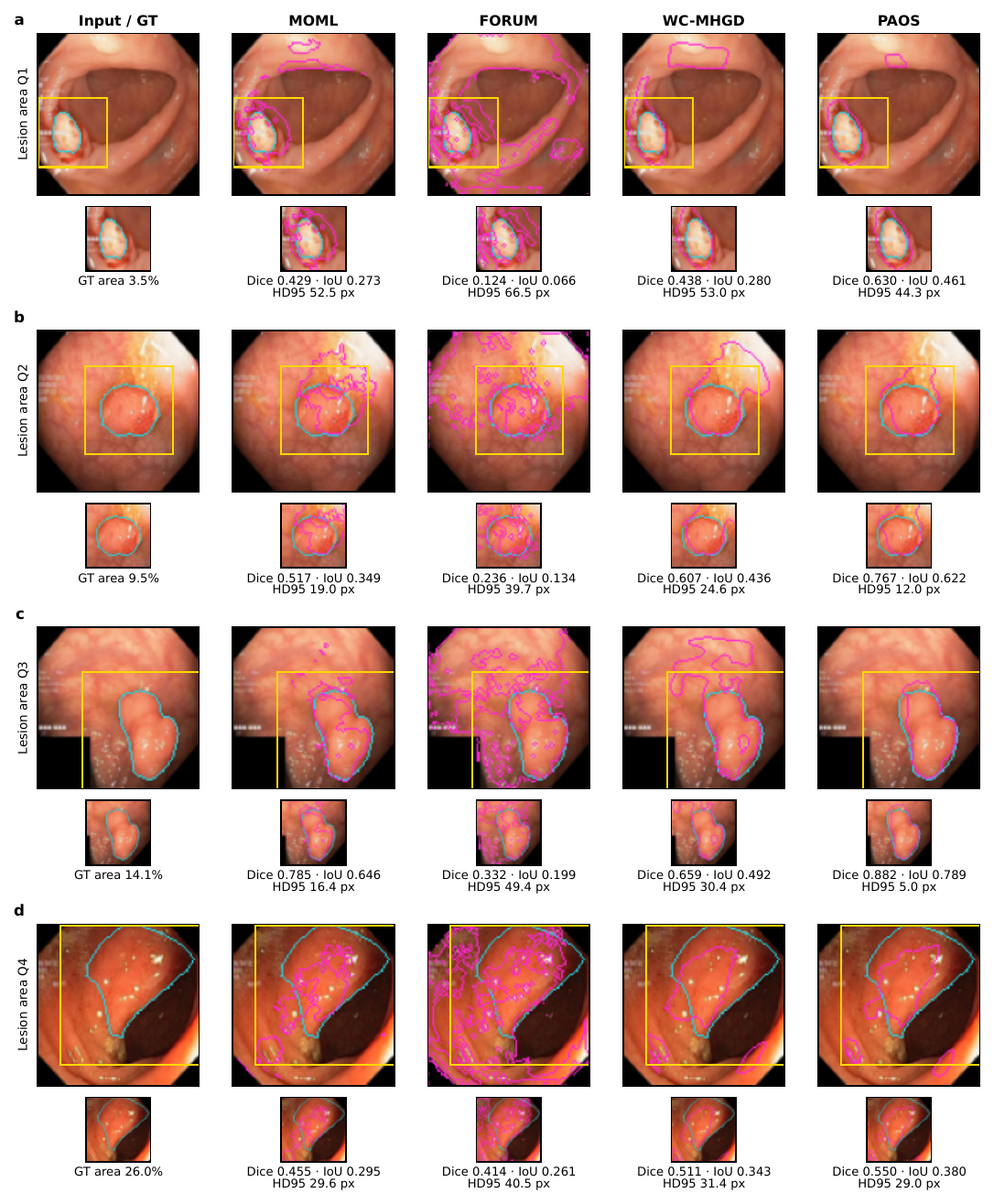}
		\vspace{-0.6cm}
	\caption{Kvasir-SEG qualitative comparisons across four lesion-area
		quartiles. All methods use the same image, region of interest, and display
		scale within each row.}
	\label{fig:kvasir_quartiles}
\end{figure}

\subsection{Additional PINN Comparison}
\label{app:pinn_momeha_comparison}

\begin{revision}
Table~\ref{tab:pinn_additional_methods} extends the heat-equation PINN
comparison to WC-Penalty~\citep{zhang2026tale} and
MOMEHA~\citep{jiang2026efficient}. PAOS performs best on test relative $L_2$
error, relative-$L_2$ AUC, and final-20 error. WC-Penalty's convex-lower-level
penalty reformulation is not designed to correct the path-dependent responses
formed by nonconvex PINN training. MOMEHA accommodates nonconvexity, but its
Moreau-envelope auxiliary response targets lower-level stationarity rather
than retaining the current response's Pareto decision to guide subsequent
training. PAOS carries that decision forward through a persistent selector,
repeatedly adjusting response formation as the Pareto geometry changes.
\end{revision}

\begin{table}[H]
\centering
\caption{\textcolor{RevisionBlue}{Additional heat-equation PINN method
comparison for Table~\ref{tab:pinn_main}.}}
\label{tab:pinn_additional_methods}
\small
\setlength{\tabcolsep}{6pt}
{\color{RevisionBlue}
\begin{tabular}{|l|c|c|c|}
\hline
Metric & MOMEHA & WC-Penalty & PAOS \\
\hline
Test rel. $L_2$ $\downarrow$
& $0.045 \pm 0.008$
& $0.247 \pm 0.039$
& $\mathbf{0.028 \pm 0.000}$ \\
Rel.-$L_2$ AUC $\downarrow$
& $0.119 \pm 0.002$
& $0.369 \pm 0.054$
& $\mathbf{0.082 \pm 0.005}$ \\
Final-20 rel. $L_2$ $\downarrow$
& $0.056 \pm 0.008$
& $0.265 \pm 0.045$
& $\mathbf{0.040 \pm 0.003}$ \\
\hline
\end{tabular}
}
\end{table}

\fi

\let\PAOSIncludeMain\undefined
\FloatBarrier

\section{Conclusion}
\label{sec:conclusion}

PAOS closes the Pareto-response loop by turning each preference-conditioned
mixture into a stateful selector that shapes the next response and its objective
geometry. Double-well and PINN ablations isolate feedback gains beyond static
selection, while four application domains demonstrate consistent gains from
feedback-guided response formation. These
results establish response formation as an actionable decision variable in
multi-objective bilevel learning. 
\textbf{Limitation and future work. } PAOS is developed for differentiable MOBL with a scalar lower-level objective
and a simplex-valued Pareto coordinator. Extending closed-loop response
selection to genuinely multi-objective lower levels and black-box evolutionary
settings is a natural next step~\citep{chauhan2026review}.

\subsection*{AI use statement}
Generative AI tools were used to assist with code development and debugging,
theoretical and experimental analysis, literature organization,
language editing, and manuscript preparation. The authors verified and take
responsibility for all technical content. 

\subsection*{Reproducibility statement}
The paper and appendix specify the update equations, task definitions,
optimization settings, preference protocols, evaluation metrics, and seed
aggregation procedures.

\bibliography{paos_iclr2027_refs}
\bibliographystyle{iclr2027_conference}

\clearpage
{\color{black}\section*{Appendix Outline}}
\begingroup
\small
\setlength{\parindent}{0pt}
\newcommand{\appoutlineentry}[3]{%
  \noindent\textcolor{black}{\textbf{#1}\enspace
  \hyperref[#2]{\textbf{#3}}}\dotfill\textbf{\pageref{#2}}\par\vspace{0.45em}}
\newcommand{\appoutlinesubentry}[3]{%
  \noindent\hspace*{1.55em}\textcolor{black}{#1\enspace
  \hyperref[#2]{#3}}\dotfill\pageref{#2}\par\vspace{0.18em}}

\appoutlineentry{A}{app:theory}{Theoretical Analysis of PAOS}
\appoutlinesubentry{A.1}{app:assumptions}{Definitions and Assumptions}
\appoutlinesubentry{A.2}{app:optimistic_proof}{Proof of Optimistic-Selection Consistency}
\appoutlinesubentry{A.3}{app:local_perturbation}{Local Response and Geometry Perturbation}
\appoutlinesubentry{A.4}{app:hg_error}{Hypergradient Approximation Error}
\appoutlinesubentry{A.5}{app:stationarity_proof}{Proof of the Pareto-Stationarity Bound}

\appoutlineentry{B}{app:double_well}{Details of the Analytical Mechanism Study}

\appoutlineentry{C}{app:implementation}{Additional Algorithmic and Complexity Details}
\appoutlinesubentry{C.1}{app:selector_schedule}{Selector Initialization and Selection Schedule}
\appoutlinesubentry{C.2}{app:warm_starts}{Finite-Step Solvers and Cost}
\appoutlinesubentry{C.3}{app:special_cases}{Ablation Configurations}

\appoutlineentry{D}{app:experimental_details}{Experimental Details}
\appoutlinesubentry{D.1}{app:common_protocol}{Preferences and Implementation}
\appoutlinesubentry{D.2}{app:mnist_hypercleaning}{MNIST Data Hyper-Cleaning}
\appoutlinesubentry{D.3}{app:fc100}{Few-shot Bilevel Learning on FC-100}
\appoutlinesubentry{D.4}{app:pinn}{Physics-Informed Learning on the Heat Equation}
\appoutlinesubentry{D.5}{app:kvasir_definition}{Kvasir-SEG Uniform-Preference Loss Design}

\appoutlineentry{E}{app:additional_results}{Additional Experimental Results}
\appoutlinesubentry{E.1}{app:quality_efficiency}{Quality-to-Target Efficiency}
\appoutlinesubentry{E.2}{app:kvasir_additional}{Additional Kvasir-SEG Results}
\appoutlinesubentry{E.3}{app:pinn_momeha_comparison}{Additional PINN Comparison}
\endgroup
\clearpage

\appendix

\section{Theoretical Analysis of PAOS}
\label{app:theory}

This appendix presents the assumptions and proofs supporting
Section~\ref{sec:theory_main}. Optimistic-selection consistency allows
non-convex lower objectives and non-unique minimizers. The hypergradient
analysis and the Pareto-stationarity bound use a regular local response
branch.

\subsection{Definitions and Assumptions}
\label{app:assumptions}
\label{app:wc_descent}

For the scalar lower-level objective in Eq.~\eqref{eq:mobl}, fix
$\balpha\in\simplex$ and define
\begin{equation}
\begin{aligned}
    \widetilde g_\rho(\bx,\by;\balpha)
    &:=g(\bx,\by)+\rho\psi_{\balpha}(\bx,\by)
    =g(\bx,\by)+\rho\sum_{s=1}^{S}\alpha_s f_s(\bx,\by),\\[-1pt]
    \psi_{\balpha}(\bx,\by)
    &:=\langle\balpha,\bF(\bx,\by)\rangle.
\end{aligned}
\end{equation}
The main-text objective $\widetilde g_t$ in
Eq.~\eqref{eq:selected_lower} is the time-varying specialization
$\widetilde g_{\rho_t}(\cdot,\cdot;\balpha_t)$.
Let
\begin{equation}
    \cY^\star(\bx):=\argmin_{\by\in\cY}g(\bx,\by),
    \qquad
    \cY^{\mathrm{opt}}_{\balpha}(\bx)
    :=\argmin_{\by\in\cY^\star(\bx)}\psi_{\balpha}(\bx,\by).
\end{equation}

For the local smooth analysis, consider an interior local-minimizer branch
$\by_\rho(\bx,\balpha)$ satisfying
$\nabla_{\by}\widetilde g_\rho(\bx,\by_\rho;\balpha)=0$
and the positive Hessian condition below. At fixed $(\bx,\balpha)$, define
\begin{align}
    \bPhi_\rho&:=\bF(\bx,\by_\rho(\bx,\balpha)),\label{eq:phi_rho_app}\\
    A_\rho&:=\nabla_{\by\by}^2\widetilde g_\rho,
    \qquad B_\rho:=\nabla_{\bx\by}^2\widetilde g_\rho,\label{eq:ab_rho_app}\\
    \bh_{\rho,s}&:=\nabla_{\bx}f_s-B_\rho A_\rho^{-1}\nabla_{\by}f_s,
    \qquad \bH_\rho:=[\bh_{\rho,1},\ldots,\bh_{\rho,S}].\label{eq:h_rho_app}
\end{align}
Here the dependence on $(\bx,\balpha)$ is suppressed; all derivatives are
evaluated at $(\bx,\by_\rho(\bx,\balpha))$ with selector $\balpha$. Let
$\blambda(\bx,\balpha)\in\simplex$ denote \rev{an exact solution of
Eq.~\eqref{eq:wc_qp} evaluated with $(\bPhi_\rho,\bH_\rho)$ at fixed
$(\rho,\balpha)$}, and define its exact direction by
$\bar\bd(\bx,\balpha):=
\bH_\rho(\bx,\balpha)
(\br\odot\blambda(\bx,\balpha))$.

The following assumptions are used selectively; Proposition
\ref{prop:optimistic_consistency} requires only
Assumption~\ref{assump:existence}, whereas the local implicit analysis uses
Assumptions~\ref{assump:local_regularity}--\ref{assump:smoothness}.

\begin{assumption}[Existence and compactness]
\label{assump:existence}
For every $\bx$ in the region of interest, $g(\bx,\cdot)$ and
$\psi_{\balpha}(\bx,\cdot)$ are continuous on $\cY$. Both $\cY^\star(\bx)$
and the minimizer set of $\widetilde g_\rho(\bx,\cdot;\balpha)$ are
nonempty for sufficiently small $\rho>0$. The surrogate minimizers lie in
a common compact lower-level sublevel set contained in $\cY$.
\end{assumption}

\begin{assumption}[Local response regularity]
\label{assump:local_regularity}
Along the selected branch visited by PAOS,
$\widetilde g_\rho(\bx,\cdot;\balpha)$ is locally $\mu$-strongly convex for
some $\mu>0$.  Equivalently, its Hessian satisfies
$\nabla_{\by\by}^2\widetilde g_\rho\succeq\mu I$ in a neighborhood containing
the iterates and the exact selected response.
\end{assumption}

\begin{assumption}[Smoothness and bounded derivatives]
\label{assump:smoothness}
The functions $g$ and $f_s$ are three-times continuously differentiable in the
local response region. Their gradients, \rev{mixed derivative matrices}, and lower-level
Hessians are Lipschitz, and the required derivatives are uniformly bounded.
The upper iterates remain in a compact subset of $\cX$.
\end{assumption}

\begin{assumption}[Conditional base-coordinator descent]
\label{assump:pareto_descent}
On an unconstrained upper domain $\cX=\R^p$, fix $\rho$ and a smooth
selected-response branch. Assume that there exists a merit function
$\mathcal M(\bx,\balpha)\geq\mathcal M_{\inf}>-\infty$ uniformly over the
visited region. Let $\E_t$ condition on the history before computing the
iteration-$t$ direction. Write the implemented direction as
$\bd_t=\bar\bd_t+\be_t+\bxi_t$, where
$\be_t:=\E_t[\bd_t]-\bar\bd_t$ is the total conditional bias and
$\E_t[\bxi_t]=0$, $\E_t\|\bxi_t\|^2\leq\sigma^2$.  There exist constants
$c_d,c_e,c_\sigma,L_\alpha>0$ such that, for a sufficiently small $\eta_x$,
\begin{align}
    \E_t[\mathcal{M}(\bx_{t+1},\balpha_t)]
    &\leq
    \mathcal{M}(\bx_t,\balpha_t)
    -c_d\eta_x\|\bar\bd_t\|^2
    +c_e\eta_x\|\be_t\|^2
    +c_\sigma\eta_x^2\sigma^2,
    \label{eq:base_descent}\\
    |\mathcal{M}(\bx,\balpha)-\mathcal{M}(\bx,\balpha')|
    &\leq
    L_\alpha\|\balpha-\balpha'\|.
    \label{eq:merit_alpha_lip}
\end{align}
Here $\bx_{t+1}=\bx_t-\eta_x\bd_t$.
Equation~\eqref{eq:wc_exact_merit_descent} establishes deterministic
fixed-selector descent for the exact weighted-Chebyshev coordinator.
Equation~\eqref{eq:base_descent} specifies the conditional descent
assumption for biased and stochastic directions, and
Eq.~\eqref{eq:merit_alpha_lip} bounds selector-induced drift.
The theorem analyzes a fixed-$\rho$ phase initialized after any finite warm-up.
\end{assumption}

\begin{revision}
\paragraph{Exact weighted-Chebyshev descent.}
Fix $\rho,\balpha$ on a smooth response branch and set
$c_s=r_s\Phi_{\rho,s}$, $P=\bH_\rho\operatorname{diag}(\br)$, so that
$P_s=\nabla_{\bx}c_s$.
Let $\blambda^\star$ minimize
$\|P\blambda\|^2-u c^\top\blambda$ over $\simplex$ and
$\bar\bd=P\blambda^\star$.
Suppose $\cX=\R^p$, $u\geq0$, and each $c_s$ has an $L$-Lipschitz gradient
along the update segment within this branch.
For $0<\eta_x\leq\min\{1/L,2/u\}$, omitting $2/u$ when $u=0$,
the merit $\mathcal M(\bx,\balpha)=\max_s c_s(\bx,\balpha)$ satisfies
\begin{equation}
\label{eq:wc_exact_merit_descent}
    \mathcal M(\bx-\eta_x\bar\bd,\balpha)
    \leq\mathcal M(\bx,\balpha)-\frac{\eta_x}{2}\|\bar\bd\|^2.
\end{equation}
QP optimality against each simplex vertex gives
$P_s^\top\bar\bd\geq\|\bar\bd\|^2+\tfrac{u}{2}(c_s-c^\top\blambda^\star)$.
By smoothness,
\[
c_s(\bx-\eta_x\bar\bd,\balpha)
\leq (1-\eta_xu/2)c_s+(\eta_xu/2)c^\top\blambda^\star
     -\eta_x(1-L\eta_x/2)\|\bar\bd\|^2
\leq \mathcal M(\bx,\balpha)-\tfrac{\eta_x}{2}\|\bar\bd\|^2.
\]
Taking the maximum over $s$ yields Eq.~\eqref{eq:wc_exact_merit_descent}.
\end{revision}

\subsection{Proof of Optimistic-Selection Consistency}
\label{app:optimistic_proof}

\begin{proof}[Proof of Proposition~\ref{prop:optimistic_consistency}]
Fix $\bx$ and $\balpha$. Write $g(\by)=g(\bx,\by)$,
$\psi(\by)=\psi_{\balpha}(\bx,\by)$, and $g^\star=\min_{\by\in\cY} g(\by)$.
Let $\by_k$ be exact surrogate minimizers along a subsequence
$\rho_k\downarrow0$ with $\by_k\to\bar\by$. The compact sublevel set in
Assumption~\ref{assump:existence} gives $\bar\by\in\cY$.
For any $\by^\star\in\cY^\star(\bx)$, surrogate optimality gives
\begin{equation}
\label{eq:optimistic_optimality}
    0\leq g(\by_k)-g^\star
    \leq \rho_k\bigl(\psi(\by^\star)-\psi(\by_k)\bigr).
\end{equation}
Continuity gives bounded $\psi(\by_k)$ and $g(\bar\by)=g^\star$.
The same inequality gives $\psi(\by_k)\leq\psi(\by^\star)$, so
$\psi(\bar\by)\leq\psi(\by^\star)$ for every $\by^\star\in\cY^\star(\bx)$.
Thus $\bar\by\in\cY^{\mathrm{opt}}_{\balpha}(\bx)$.
\end{proof}

At fixed $\bx$, positive linear weights in Eq.~\eqref{eq:ideal_optimistic}
select supported Pareto-efficient responses in $\bF(\bx,\cY^\star(\bx))$,
which may also contain unsupported efficient responses.

\subsection{Local Response and Geometry Perturbation}
\label{app:local_perturbation}

\begin{proof}[Proof of Proposition~\ref{prop:response_shift}]
Fix $\bx$ and define
\begin{equation}
    \mathcal{G}(\by,\rho)
    :=
    \nabla_{\by}g(\bx,\by)
    +\rho\nabla_{\by}\psi_{\balpha}(\bx,\by).
\end{equation}
At $(\by^0,0)$, we have
$\mathcal{G}(\by^0,0)=0$ and
$\nabla_{\by}\mathcal{G}(\by^0,0)=G\succ0$.
The implicit function theorem therefore gives a differentiable branch
$\by_{\rho,\balpha}$ satisfying
$\mathcal{G}(\by_{\rho,\balpha},\rho)=0$ for sufficiently small $\rho$.
Differentiating at $\rho=0$ gives
\begin{equation}
    G
    \left.
    \frac{\mathrm{d}\by_{\rho,\balpha}}{\mathrm{d}\rho}
    \right|_{\rho=0}
    +\ba_{\balpha}
    =0,
\end{equation}
and hence
\begin{equation}
    \left.
    \frac{\mathrm{d}\by_{\rho,\balpha}}{\mathrm{d}\rho}
    \right|_{\rho=0}
    =-G^{-1}\ba_{\balpha}.
\end{equation}
Taylor expansion proves Eq.~\eqref{eq:response_shift}.

Using
$\Delta\by:=\by_{\rho,\balpha}-\by^0
=-\rho G^{-1}\ba_{\balpha}+\mathcal{O}(\rho^2)$,
we obtain
\begin{align}
    \psi_{\balpha}(\bx,\by_{\rho,\balpha})
    -\psi_{\balpha}(\bx,\by^0)
    &=
    \ba_{\balpha}^\top\Delta\by
    +\mathcal{O}(\|\Delta\by\|^2)\\
    &=
    -\rho\ba_{\balpha}^\top G^{-1}\ba_{\balpha}
    +\mathcal{O}(\rho^2),
\end{align}
which proves the first objective expansion in
Proposition~\ref{prop:response_shift}. Since
$\nabla_{\by}g(\bx,\by^0)=0$,
\begin{align}
    g(\bx,\by_{\rho,\balpha})-g(\bx,\by^0)
    &=
    \frac{1}{2}\Delta\by^\top G\Delta\by
    +\mathcal{O}(\|\Delta\by\|^3)\\
    &=
    \frac{\rho^2}{2}
    \ba_{\balpha}^\top G^{-1}\ba_{\balpha}
    +\mathcal{O}(\rho^3),
\end{align}
proving the corresponding lower-objective expansion.
\end{proof}

Applying the same response expansion to the fixed user criterion gives
\begin{equation*}
    \psi_{\br}(\bx,\by_{\rho,\balpha})-\psi_{\br}(\bx,\by^0)
    =-\rho\,\ba_{\br}^{\top}G^{-1}\ba_{\balpha}
      +\mathcal O(\rho^2),
    \qquad
    \ba_{\br}:=\nabla_{\by}\psi_{\br}(\bx,\by^0).
\end{equation*}
A positive cross-alignment term
$\ba_{\br}^{\top}G^{-1}\ba_{\balpha}$
gives first-order improvement in the fixed user criterion $\psi_{\br}$.

\begin{corollary}[Hypergradient perturbation]
\label{cor:geometry_shift}
Under the conditions of Proposition~\ref{prop:response_shift}, if the
derivatives entering the implicit hypergradient are locally Lipschitz, then
\begin{equation}
    \|\bH_\rho(\bx,\balpha)-\bH_0(\bx)\|_{\mathrm{op}}
    \leq C_H\rho
\end{equation}
for some local constant $C_H>0$.
\end{corollary}

\begin{proof}
Equation~\eqref{eq:response_shift} and a first-order Taylor expansion give
$f_j(\bx,\by_{\rho,\balpha})-f_j(\bx,\by^0)
=-\rho\nabla_{\by}f_j(\bx,\by^0)^\top G^{-1}\ba_{\balpha}
+\mathcal O(\rho^2)$.
The hypergradient bound follows from the $\mathcal O(\rho)$ response shift
and the local Lipschitz conditions. Together with the explicit $\rho$-dependence
of $\widetilde g_\rho$, they give $\mathcal O(\rho)$ changes in the gradient,
mixed-derivative, and lower-Hessian factors. Positive definiteness along
the local continuation bounds both inverse norms. The identity
$A_\rho^{-1}-A_0^{-1}=A_\rho^{-1}(A_0-A_\rho)A_0^{-1}$ then gives an
$\mathcal O(\rho)$ change in the inverse-Hessian factor. Substitution into
Eq.~\eqref{eq:h_rho_app} yields the claimed bound.
\end{proof}

Define the standard Pareto-stationarity residual of a hypergradient matrix
$H$ by
\begin{equation}
    s(H)
    :=
    \min_{\bw\in\simplex}\|H\bw\|.
\end{equation}
Since $\|\bw\|_2\leq1$ for $\bw\in\simplex$,
Corollary~\ref{cor:geometry_shift} also gives
\begin{equation}
    s(\bH_0(\bx))
    \leq
    s(\bH_\rho(\bx,\balpha))
    +C_H\rho.
\end{equation}
For $r_{\min}:=\min_s r_s>0$, let
$m=\one^\top(\br\odot\blambda^\star)\geq r_{\min}$.
The vector $(\br\odot\blambda^\star)/m$ lies in $\simplex$, so
$s(\bH_\rho)\leq\|\bar\bd\|/m$ and hence
\begin{equation}
\label{eq:original_stationarity_link}
    s(\bH_0(\bx))
    \leq\frac{\|\bar\bd(\bx,\balpha)\|}{r_{\min}}+C_H\rho.
\end{equation}
A distance bound follows under the local error bound
$\operatorname{dist}(\bx,\mathcal S_0)\leq
\kappa\,s(\bH_0(\bx))$,
where $\mathcal S_0$ denotes the stationary set of the original branch.

\subsection{Hypergradient Approximation Error}
\label{app:hg_error}

Let $\by^\star:=\by_\rho(\bx,\balpha)$ be an exact selected response.  Define
\begin{align}
    A^\star&:=
    \nabla_{\by\by}^2
    \widetilde g_\rho(\bx,\by^\star;\balpha),
    \qquad B^\star:=
    \nabla_{\bx\by}^2
    \widetilde g_\rho(\bx,\by^\star;\balpha),\\
    \bh_s^\star&:=
    \nabla_{\bx}f_s(\bx,\by^\star)
    -B^\star(A^\star)^{-1}
    \nabla_{\by}f_s(\bx,\by^\star).
\end{align}
At an approximate response $\widehat\by$, let
\begin{equation}
    \widehat A
    :=
    \nabla_{\by\by}^2
    \widetilde g_\rho(\bx,\widehat\by;\balpha),
    \qquad
    \widehat B
    :=
    \nabla_{\bx\by}^2
    \widetilde g_\rho(\bx,\widehat\by;\balpha),
\end{equation}
and suppose $\widehat\bv_s$ has damped linear residual
\begin{equation}
    \be_{v,s}
    :=
    (\widehat A+\delta I)\widehat\bv_s
    -\nabla_{\by}f_s(\bx,\widehat\by).
\end{equation}

\begin{lemma}[Hypergradient error decomposition]
\label{lem:hg_error}
Under Assumptions~\ref{assump:local_regularity} and
\ref{assump:smoothness}, let
$\epsilon_y:=\|\nabla_{\by}\widetilde g_\rho
(\bx,\widehat\by;\balpha)\|$. There exist constants
$C_y,C_v,C_\delta>0$ such that
\begin{equation}
\label{eq:hg_error_bound}
    \|\widehat\bh_s-\bh_s^\star\|
    \leq
    C_y\epsilon_y
    +C_v\|\be_{v,s}\|
    +C_\delta\delta.
\end{equation}
\end{lemma}

\begin{proof}
Local $\mu$-strong convexity implies
\begin{equation}
    \|\widehat\by-\by^\star\|
    \leq
    \mu^{-1}\epsilon_y.
\end{equation}
The inverse perturbation identity and the derivative Lipschitz bounds therefore
control the error caused by replacing $(A^\star,B^\star,\by^\star)$ with
$(\widehat A,\widehat B,\widehat\by)$ by
$\mathcal{O}(\epsilon_y)$.  Next,
\begin{equation}
    \left\|(\widehat A+\delta I)^{-1}
    -(\widehat A)^{-1}\right\|_{\mathrm{op}}
    \leq
    \frac{\delta}{\mu(\mu+\delta)}
    \leq
    \frac{\delta}{\mu^2},
\end{equation}
which gives the damping term.  Finally,
\begin{equation}
    \left\|
    \widehat\bv_s
    -(\widehat A+\delta I)^{-1}
    \nabla_{\by}f_s(\bx,\widehat\by)
    \right\|
    \leq
    \frac{1}{\mu+\delta}\|\be_{v,s}\|.
\end{equation}
Multiplying by the uniformly bounded \rev{mixed derivative matrix} and combining the three
terms proves Eq.~\eqref{eq:hg_error_bound}.
\end{proof}

\begin{revision}
\paragraph{QP gap and direction accuracy.}
Write the exact coordination subproblem as
$Q(\blambda)=\|P\blambda\|^2-u c^\top\blambda$, with
$P=\bH_\rho\operatorname{diag}(\br)$ and $c=\br\odot\bPhi_\rho$.
Let $\widehat Q$ use approximate inputs $(\widehat P,\widehat c)$, and let
$\widehat\blambda\in\simplex$ have $\widehat Q$-gap at most
$\varepsilon_{\mathrm{QP}}$.
For an exact minimizer $\blambda^\star$, set
$\widehat\bd=\widehat P\widehat\blambda$, $\bar\bd=P\blambda^\star$, and
\[
    \Delta_Q=(\|\widehat P\|_{\mathrm{op}}+\|P\|_{\mathrm{op}})
    \|\widehat P-P\|_{\mathrm{op}}+u\|\widehat c-c\|_\infty.
\]
Then
\begin{equation}
\label{eq:qp_direction_stability}
    \|\widehat\bd-\bar\bd\|
    \leq\|\widehat P-P\|_{\mathrm{op}}
       +\sqrt{2\Delta_Q+\varepsilon_{\mathrm{QP}}}.
\end{equation}
The input bound $\sup_{\simplex}|\widehat Q-Q|\leq\Delta_Q$ gives
$Q(\widehat\blambda)-Q(\blambda^\star)\leq
2\Delta_Q+\varepsilon_{\mathrm{QP}}$. QP optimality bounds
$\|P(\widehat\blambda-\blambda^\star)\|^2$ by this exact gap.
Adding $\|(\widehat P-P)\widehat\blambda\|$ proves the result.
With zero input error and gap, the bound also gives uniqueness of the exact
direction.

Lemma~\ref{lem:hg_error} controls the matrix input error. The response-induced
value error is controlled by local Lipschitz continuity and
$\|\widehat\by-\by^\star\|\leq\epsilon_y/\mu$.
Jensen's inequality gives
$\|\be_t\|^2\leq\E_t\|\bd_t-\bar\bd_t\|^2$ for the implemented direction,
so the bound applies to the conditional bias through the input errors
(including sampling error) and the QP gap.
\end{revision}

\subsection{Proof of the Pareto-Stationarity Bound}
\label{app:stationarity_proof}

Reindex the fixed-$\rho$ phase after warm-up from $t=0$. Let
$\bar\bd_t:=\bar\bd(\bx_t,\balpha_t)$ be the exact coordination direction
defined in Appendix~\ref{app:assumptions}. Define the cumulative selector
drift $V_N:=\sum_{t=0}^{N-1}\E\|\balpha_{t+1}-\balpha_t\|$ and the mean
squared direction bias $\overline e_N^2:=N^{-1}\sum_{t=0}^{N-1}\E\|\be_t\|^2$,
where $\be_t:=\E_t[\bd_t]-\bar\bd_t$.

\begin{lemma}[Finite-iteration bound on the selected direction]
\label{lem:selected_direction_bound}
Under Assumption~\ref{assump:pareto_descent},
\[
    \frac{1}{N}\sum_{t=0}^{N-1}\E\|\bar\bd_t\|^2
    \leq \mathcal E_N
    :=
    \frac{C_0+C_1 V_N}{\eta_x N}
    + C_2\overline e_N^2 + C_3\eta_x\sigma^2,
\]
where $C_0=(\E\mathcal M(\bx_0,\balpha_0)-\mathcal M_{\inf})/c_d$,
$C_1=L_\alpha/c_d$, $C_2=c_e/c_d$, and $C_3=c_\sigma/c_d$.
\end{lemma}

\begin{proof}
Condition on the history at $t$, apply Eq.~\eqref{eq:base_descent}, and account
for the selector change using Eq.~\eqref{eq:merit_alpha_lip}. Taking total
expectations gives
\begin{align}
    \E\mathcal{M}(\bx_{t+1},\balpha_{t+1})
    &\leq
    \E\mathcal{M}(\bx_{t+1},\balpha_t)
    +L_\alpha
    \E\|\balpha_{t+1}-\balpha_t\|\\
    &\leq
    \E\mathcal{M}(\bx_t,\balpha_t)
    -c_d\eta_x\E\|\bar\bd_t\|^2\nonumber\\[-1pt]
    &\quad
    +c_e\eta_x\E\|\be_t\|^2
    +c_\sigma\eta_x^2\sigma^2\nonumber\\[-1pt]
    &\quad
    +L_\alpha
    \E\|\balpha_{t+1}-\balpha_t\|.
\end{align}
Summing from $t=0$ to $N-1$ and using the lower bound
$\mathcal{M}_{\inf}$ yields
\begin{equation}
    c_d\eta_x
    \sum_{t=0}^{N-1}\E\|\bar\bd_t\|^2
    \leq
    \E\mathcal{M}(\bx_0,\balpha_0)-\mathcal{M}_{\inf}
    +L_\alpha V_N
    +c_e\eta_x
    \sum_{t=0}^{N-1}\E\|\be_t\|^2
    +c_\sigma N\eta_x^2\sigma^2.
\end{equation}
Dividing by $c_d\eta_x N$ gives the stated bound.
\end{proof}

\begin{proof}[Proof of Theorem~\ref{thm:variation_stationarity}]
The original-branch condition identifies $\bH_0(\bx_t)$ with the
hypergradients of the original upper objectives restricted to
$\by^0(\bx_t)$. The common continuation and uniform constant $C_H$ give
Eq.~\eqref{eq:original_stationarity_link} at each iterate, with
$\bar\bd(\bx_t,\balpha_t)=\bar\bd_t$. Squaring the bound and applying
$(a+b)^2\leq 2a^2+2b^2$ gives
\[
    s(\bH_0(\bx_t))^2
    \leq \frac{2}{r_{\min}^2}\|\bar\bd_t\|^2+2C_H^2\rho^2.
\]
Taking expectations, averaging over $t$, and applying
Lemma~\ref{lem:selected_direction_bound} gives
Eq.~\eqref{eq:original_branch_average_bound}.
\end{proof}

\section{Details of the Analytical Mechanism Study}
\label{app:double_well}

\paragraph{Problem formulation.}
We use the analytical double-well problem from
Section~\ref{sec:experiments} to examine how Pareto feedback shapes
lower-level response formation.
Let the upper variable be $x\in\mathbb{R}$ and the lower variable be
$(z,v)\in\mathbb{R}^2$. Define the lower objective and two upper objectives as
\begin{align}
    g(x,z,v)&=\frac{1}{4}(z^2-1)^2+\frac{1}{2}(v-x)^2,
    \qquad b(z)=\frac{2+3z-z^3}{4},\label{eq:dw_lower_app}\\
    f_1(x,z,v)&=(v+1)^2+b(z),
    \qquad f_2(x,z,v)=(v-1)^2+1-b(z).\label{eq:dw_upper_1_app}
\end{align}
For every $x\in\mathbb{R}$, the original lower objective satisfies
$g^\star(x)=0$, and its optimal response set consists of two equally valid
branches:
\begin{equation}
    \mathcal{S}(x)
    :=\operatorname*{arg\,min}_{z,v}g(x,z,v)
    =\{(-1,x),(+1,x)\}.
\label{eq:dw_response_set}
\end{equation}

\paragraph{Attraction-basin mechanism.}
For a frozen selector $\balpha=(\alpha_1,1-\alpha_1)$, the branch derivative
of the selected objective factorizes as
\begin{equation}
    \partial_z\widetilde g
    =(z^2-1)\left[z-\frac{3\rho}{4}(2\alpha_1-1)\right],
    \qquad z_{\mathrm{sep}}=\frac{3\rho}{4}(2\alpha_1-1).
    \label{eq:dw_separatrix}
\end{equation}
When $|z_{\mathrm{sep}}|<1$, $z=\pm1$ are stable equilibria and
$z_{\mathrm{sep}}$ is the intervening unstable equilibrium of the frozen
gradient flow. Pareto feedback shifts this boundary during response
formation, allowing a near-boundary initialization to enter the preferred
basin. Since $z=\pm1$ remain stationary for every selector, feedback acts
on basin allocation before the response settles.
For $\rho=0.15$, the boundary satisfies $|z_{\mathrm{sep}}|\leq0.1125$.

\paragraph{Weighted-sum reference solution.}
Restricting the preference-weighted upper objective
$J_w=w f_1+(1-w)f_2$ to the two exact response branches gives
\begin{align}
    J_w(x,-1,x)
    &=
    \bigl(x-(1-2w)\bigr)^2
    +4w(1-w)+(1-w),                                                \label{eq:dw_left}\\
    J_w(x,+1,x)
    &=
    \bigl(x-(1-2w)\bigr)^2
    +4w(1-w)+w.                                                    \label{eq:dw_right}
\end{align}
Both branches share the same optimal upper variable
$x_w^\star=1-2w$. Comparing their remaining branch-dependent terms yields
$z_w^\star=-1$ for $w>1/2$ and $z_w^\star=+1$ for $w<1/2$, with
\begin{equation}
    J_w^\star=4w(1-w)+\min\{w,1-w\}.
\end{equation}
At $w=1/2$, the two branches have equal objective values.
We use off-center preferences to obtain a unique preferred branch.
The weighted-sum oracle provides the reference for evaluating attained
responses, while the weighted-Chebyshev subproblem in
Eq.~\eqref{eq:wc_qp} determines the optimization direction.

\paragraph{Experimental protocol.}
We consider $w\in\{0.1,0.3,0.7,0.9\}$. For each preference, the 20 paired
initializations have ten starts on either side of the branch boundary,
with $x_0=v_0=0$ and $|z_0|\in[0.01,0.08]$.
All methods use the same paired initial states.

Each run uses 80 outer iterations and five lower-level gradient steps per outer
iteration. The lower and upper learning rates are $0.1$ and $0.04$,
respectively, and implicit-gradient methods use five conjugate-gradient
iterations with damping $10^{-3}$. PAOS uses $\rho=0.15$, feedback rate
$\beta=1$, and $u=10$ in the weighted-Chebyshev subproblem.
The balanced study uses seeds 20--39.
WC-MHGD is the matched open-loop counterpart of PAOS.
WS-HG and TCH-HG apply the same preference through weighted-sum and
Chebyshev hypergradient updates, respectively, with lower-level training
governed by the original objective.

\paragraph{Evaluation metrics.}
A terminal response is preference-consistent when
$\operatorname{sign}(z_T)=-\operatorname{sign}(w-\tfrac12)$,
$|z_T|\geq0.9$, and $g(x_T,z_T,v_T)\leq10^{-3}$.
For evaluation, set
$\bar z_t\in\operatorname*{arg\,min}_{s\in\{-1,+1\}}|z_t-s|$
and $\bar v_t=x_t$. The projected weighted-oracle regret is
$\mathcal{R}^{(t)}_w=J_w(x_t,\bar z_t,\bar v_t)-J_w^\star$.
Optimization uses the unprojected trajectory.

\paragraph{Preference-wise branch selection.}
Table~\ref{tab:dw_branch_results} reports preference-consistent branch
counts for each tested preference.
PAOS reaches the preferred valid response in all 20 paired runs per
preference, while each open-loop control succeeds in 10.

\begin{table}[H]
    \centering
    \caption{Preference-consistent response counts over 20 paired
    initializations per preference. Bold denotes the best result.}
    \label{tab:dw_branch_results}
    \small
    \setlength{\tabcolsep}{5pt}
    \begin{tabular}{|l|c|c|c|c|}
        \hline
        Method
        & $w=0.1$
        & $w=0.3$
        & $w=0.7$
        & $w=0.9$\\ \hline
        WS-HG   & $10/20$ & $10/20$ & $10/20$ & $10/20$\\
        TCH-HG  & $10/20$ & $10/20$ & $10/20$ & $10/20$\\
        WC-MHGD & $10/20$ & $10/20$ & $10/20$ & $10/20$\\
        PAOS    & $\mathbf{20/20}$ & $\mathbf{20/20}$
                & $\mathbf{20/20}$ & $\mathbf{20/20}$\\
        \hline
    \end{tabular}
\end{table}

At $\rho=0.15$, PAOS obtains valid preference-compatible responses in
all $80/80$ runs, with a median terminal lower-level gap of
$2.32\times10^{-6}$.
A separate selection-strength study uses seeds 0--19 and 20 near-boundary
starts per preference under the same 80-step configuration.
At $\rho=0.15$, it also yields $80/80$ valid preference-compatible
responses, with a median lower-level gap of $2.4\times10^{-6}$.

At $w=0.9$, the median projected oracle regrets over the 20 balanced
starts are $0.147$, $0.492$, $0.400$, and $0.463$ for PAOS, WC-MHGD,
WS-HG, and TCH-HG, respectively.
Projected regret measures upper-objective quality relative to the
weighted-sum reference, complementing the branch-selection counts.

\paragraph{Feedback-strength ablation.}
Table~\ref{tab:dw_beta_ablation} varies only $\beta$ to isolate dynamic
feedback. Static selection reaches the preferred sign in $67/80$ runs
but satisfies the valid-response criterion in none.
With $\beta=1$, PAOS reaches a valid preferred response in all $80/80$
runs and reduces median regret from $0.243$ to $0.134$.

\begin{table}[H]
    \centering
    \caption{Double-well feedback-rate ablation over 80 paired near-boundary
    runs; valid branch requires the preferred sign, $|z_T|\geq0.9$,
    and $g(x_T,z_T,v_T)\leq10^{-3}$.}
    \label{tab:dw_beta_ablation}
    \small
    \setlength{\tabcolsep}{4.5pt}
    \begin{tabular}{|l|c|c|c|c|c|}
        \hline
        Metric & $\beta=0$ (static) & $\beta=0.25$ & $\beta=0.50$
        & $\beta=0.70$ & $\beta=1.00$ (PAOS) \\ \hline
        Preferred sign $\uparrow$ & $67/80$ & $76/80$ & $78/80$
        & $79/80$ & $\mathbf{80/80}$ \\
        Valid branch $\uparrow$ & $0/80$ & $76/80$ & $78/80$
        & $79/80$ & $\mathbf{80/80}$ \\
        Median regret $\downarrow$ & $0.243$ & $0.148$ & $0.147$
        & $0.147$ & $\mathbf{0.134}$ \\
        \hline
    \end{tabular}
\end{table}

\paragraph{Initialization-distance sweep.}
At $w=0.9$, we vary positive initial $z_0$ while keeping $x_0=v_0=0$
and the preceding 80-step configuration fixed.
Using random seed 2027, we sample 20 paired starts uniformly from each
interval in Table~\ref{tab:dw_basin}.
The transition in branch-selection counts across intervals is consistent
with the bounded attraction-boundary displacement in
Eq.~\eqref{eq:dw_separatrix}.
\FloatBarrier
\begin{table*}[t]
    \centering
    \caption{Valid preferred-branch counts versus initial distance at $w=0.9$
    (20 paired starts per interval, $\rho=0.15$). Initial $z_0$ intervals
    are shown in units of $10^{-3}$.}
    \label{tab:dw_basin}
    \scriptsize
    \setlength{\tabcolsep}{3pt}
    \resizebox{\textwidth}{!}{%
    \begin{tabular}{|l|*{11}{c|}}
    \hline
    Method & $[2,15]$ & $[15,30]$ & $[30,45]$ & $[45,60]$
    & $[60,75]$ & $[75,90]$ & $[90,120]$ & $[120,180]$
    & $[180,350]$ & $[350,600]$ & $[600,850]$\\ \hline
    PAOS & $20/20$ & $20/20$ & $20/20$ & $20/20$
    & $20/20$ & $20/20$ & $9/20$ & $0/20$
    & $0/20$ & $0/20$ & $0/20$\\
    WC-MHGD & $0/20$ & $0/20$ & $0/20$ & $0/20$
    & $0/20$ & $0/20$ & $0/20$ & $0/20$
    & $0/20$ & $0/20$ & $0/20$\\ \hline
    \end{tabular}
    }
\end{table*}

\paragraph{Separate aligned-objective control.}
\label{app:dw_aligned}
Figure~\ref{fig:double_well_mechanism}(c) and Figure~\ref{fig:motivation}(b)
use a distinct control problem that retains $g$ and $b$ but aligns the
continuous upper-objective terms:
\begin{equation}
    f_1^{\mathrm{align}}=v^2+b(z),\qquad
    f_2^{\mathrm{align}}=v^2+1-b(z).
\end{equation}
Its exact preferred response has $x_w^\star=v_w^\star=0$, branch
$z_w^\star$ as above, and weighted-sum optimum $\min\{w,1-w\}$.
Projected regret is
$x_t^2+|2w-1|\,\mathbf1\{\operatorname{sign}(z_t)\neq z_w^\star\}$.
We use $w\in\{0.1,0.9\}$, seeds 0--11 for each preference,
$x_0=v_0=1.5$, and balanced signs with $|z_0|\in[0.01,0.08]$.
The displayed hard-start subset contains the six initially incompatible
signs per preference, pooled into 12 paired starts. Runs last 600 outer
iterations, with five lower steps, lower/upper learning rates $0.1/0.04$,
five CG steps, damping $10^{-3}$, $u=10$, and $\beta=1$.
PAOS anneals $\rho_t=0.2[1-(t+1)/600]$ for $t=0,\ldots,599$.
The final median projected regret is $3.16\times10^{-4}$ for PAOS and
$0.8$ for each open-loop control.
This control evaluates long-horizon refinement following branch selection
with aligned continuous upper objectives.

\section{Additional Algorithmic and Complexity Details}
\label{app:implementation}

\subsection{Selector Initialization and Selection Schedule}
\label{app:selector_schedule}

We initialize the response-selector state with the user preference,
$\balpha_0=\br$. The response-selection coefficient is increased smoothly
during an initial warm-up period:
\begin{equation}
\label{eq:rho_schedule_app}
    \rho_t
    =
    \rho_0
    +
    (\rho_{\max}-\rho_0)
    \min\!\left\{1,\frac{t+1}{W}\right\}^{p_\rho},
\end{equation}
Here $W$ is the warm-up length and $p_\rho$ controls the schedule shape.
The schedule gradually activates preference-guided response shaping.
The 80-step double-well study uses constant $\rho$, while the
aligned-objective control uses the annealing schedule in
Appendix~\ref{app:dw_aligned}.

\subsection{Finite-Step Solvers and Cost}
\label{app:warm_starts}
\label{app:oracle_accounting}

PAOS warm-starts the lower response and maintains $S$ CG states.
Detached lower steps and the implicit hypergradient in
Eq.~\eqref{eq:hypergradient} keep memory independent of the unrolling
depth $T$. An outer iteration costs approximately
$T(C_g+C_F)+SK C_{\mathrm{HVP}}+C_{\mathrm{QP}}(S)$,
where $C_g$, $C_F$, and $C_{\mathrm{HVP}}$ denote the costs of a lower
gradient, selector-weighted upper derivatives, and a Hessian-vector
product, respectively. The QP has $S$ variables; selector feedback
requires $\mathcal{O}(S)$ arithmetic and memory.

\subsection{Ablation Configurations}
\label{app:special_cases}

The following configurations isolate the components of PAOS:
\begin{center}
\small
\begin{tabular}{|p{0.25\linewidth}|p{0.20\linewidth}|p{0.43\linewidth}|}
\hline
Configuration & Setting & Interpretation \\ \hline
Open-loop baseline & $\rho_t=0$ & Original lower response \\
Uniform selection & $\balpha_t=\one/S$ & Preference-agnostic response selection \\
Static preference & $\beta=0,\ \balpha_0=\br$ & No dynamic Pareto feedback \\
Full PAOS & $\beta>0$, dynamic $\balpha_t$ & Preference-aware Pareto feedback \\
\hline
\end{tabular}
\end{center}

The main ablation uses matched lower-step and hypergradient budgets
to separate the contributions of response selection, preference
conditioning, and dynamic Pareto feedback.

\def\PAOSIncludeAppendix{1}

\let\PAOSIncludeAppendix\undefined

\end{document}